\pdfoutput=1

\documentclass[11pt]{article}
\DeclareUnicodeCharacter{0307}{\.}
\usepackage[]{acl}

\usepackage{times}
\usepackage{latexsym}

\usepackage[T1]{fontenc}

\usepackage[utf8]{inputenc}

\usepackage{microtype}
\usepackage{graphicx}
\usepackage{booktabs} 
\usepackage{tabularx}
\usepackage{multirow} 
\usepackage{hyperref}
\usepackage{lipsum}
\usepackage{xltabular}
\usepackage{longtable}
\usepackage{cals}
\usepackage{xcolor,colortbl}
\usepackage[normalem]{ulem}
\usepackage{pifont}

\def \ourbenchmark{\textsc{Sahara}}

\usepackage{soul}
\usepackage{subfigure}  
\usepackage{appendix}

\usepackage{pdflscape}
\usepackage{rotating}
\usepackage{ragged2e}

\usepackage{cuted}
\usepackage{capt-of} 

\title{\includegraphics[scale=0.04]{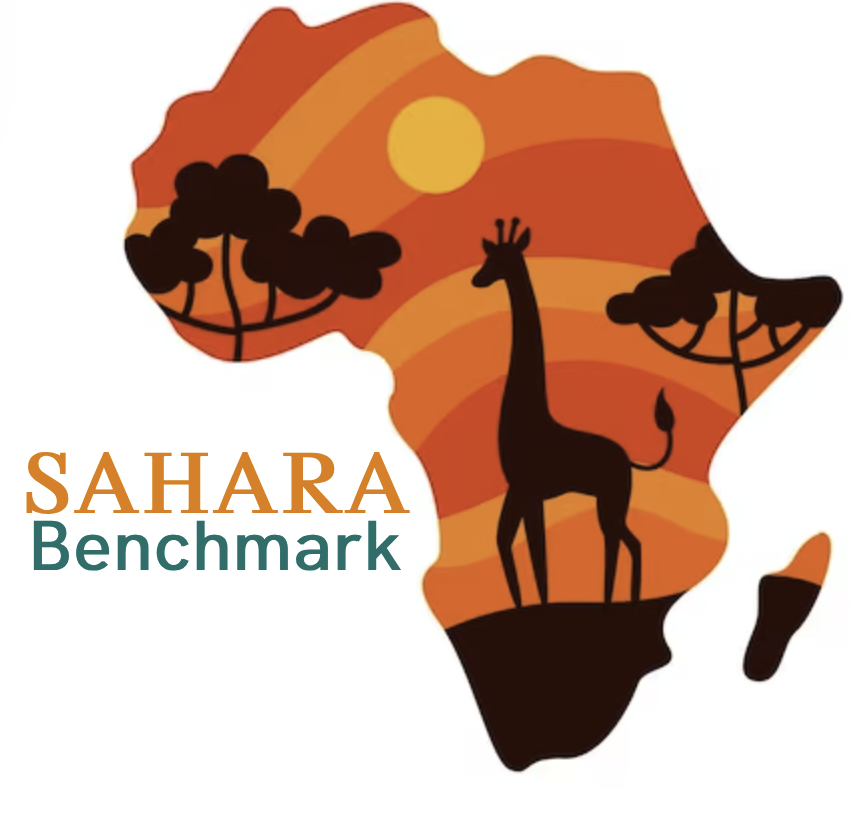} Where Are We?\\ Evaluating LLM Performance on African Languages}

\author{\begin{tabular}[c]{@{}c@{}}
\normalsize  Ife Adebara$^{\xi}$ ~  Hawau Olamide Toyin$^{\Omega}$ ~ Nahom Tesfu Ghebremichael$^{\Omega}$ \\
\normalsize  AbdelRahim Elmadany$^{\xi}$ ~ Muhammad Abdul-Mageed$^{\xi,\Omega,\lambda}$\end{tabular}\\
\normalsize $^{\xi}$The University of British Columbia ~~~~~ $^{\Omega}$MBZUAI ~~~~~ $^\lambda$ Invertible AI\\ %
  \texttt{\normalsize \{ife.adebara,a.elmadany,muhammad.mageed\}@ubc.ca}}

\begin{document}
\maketitle

\begin{strip}
\centering
\resizebox{0.93\textwidth}{!}{%
\begin{tabular}{ll}
\begin{minipage}{0.47\textwidth}
    \centering
    \includegraphics[width=\textwidth]{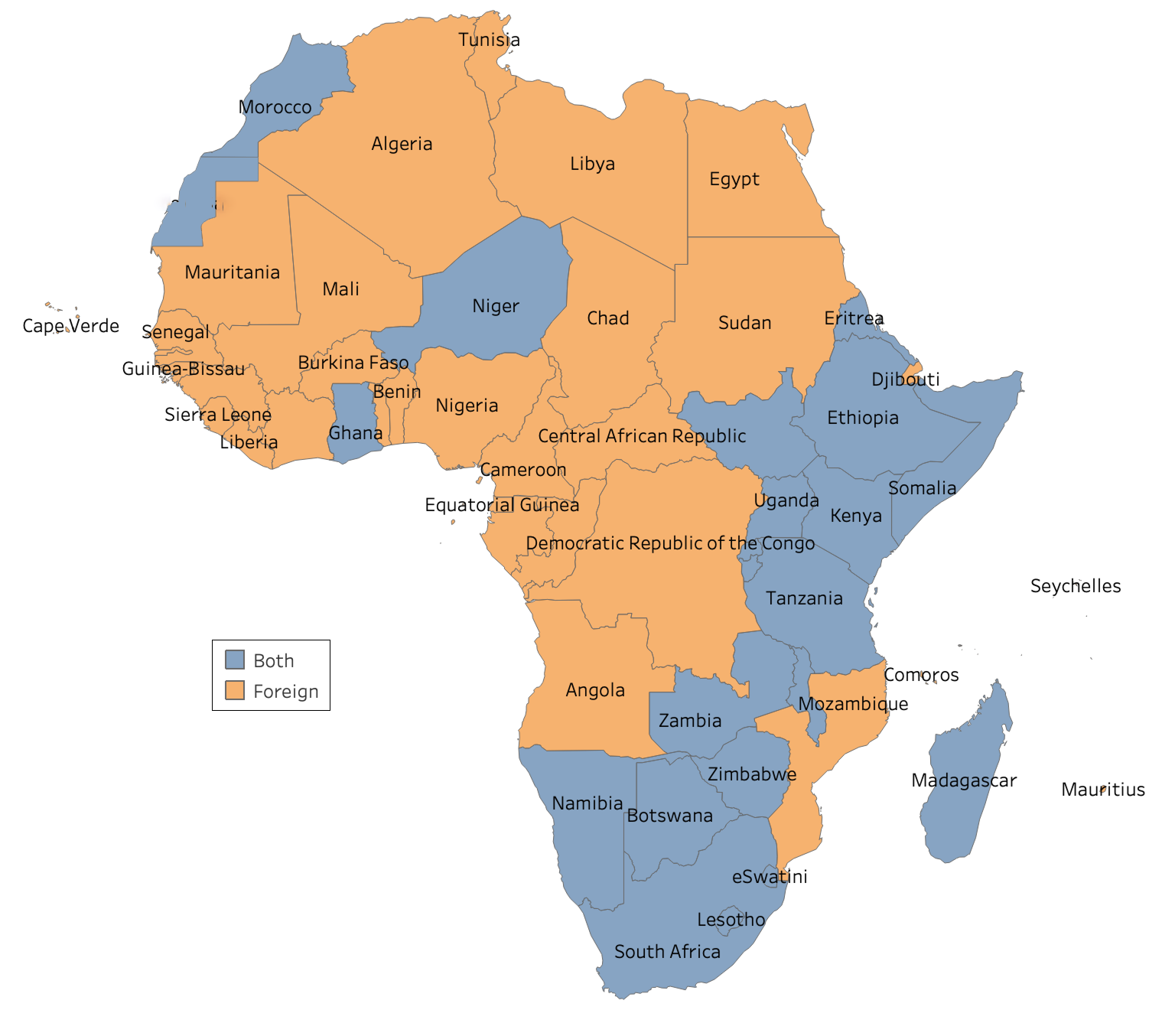}
    \justifying
    \small (a) Language policies for education across Africa.
    \label{fig:map_edu}
\end{minipage} &
\begin{minipage}{0.49\textwidth}
    \centering
    \includegraphics[width=\textwidth]{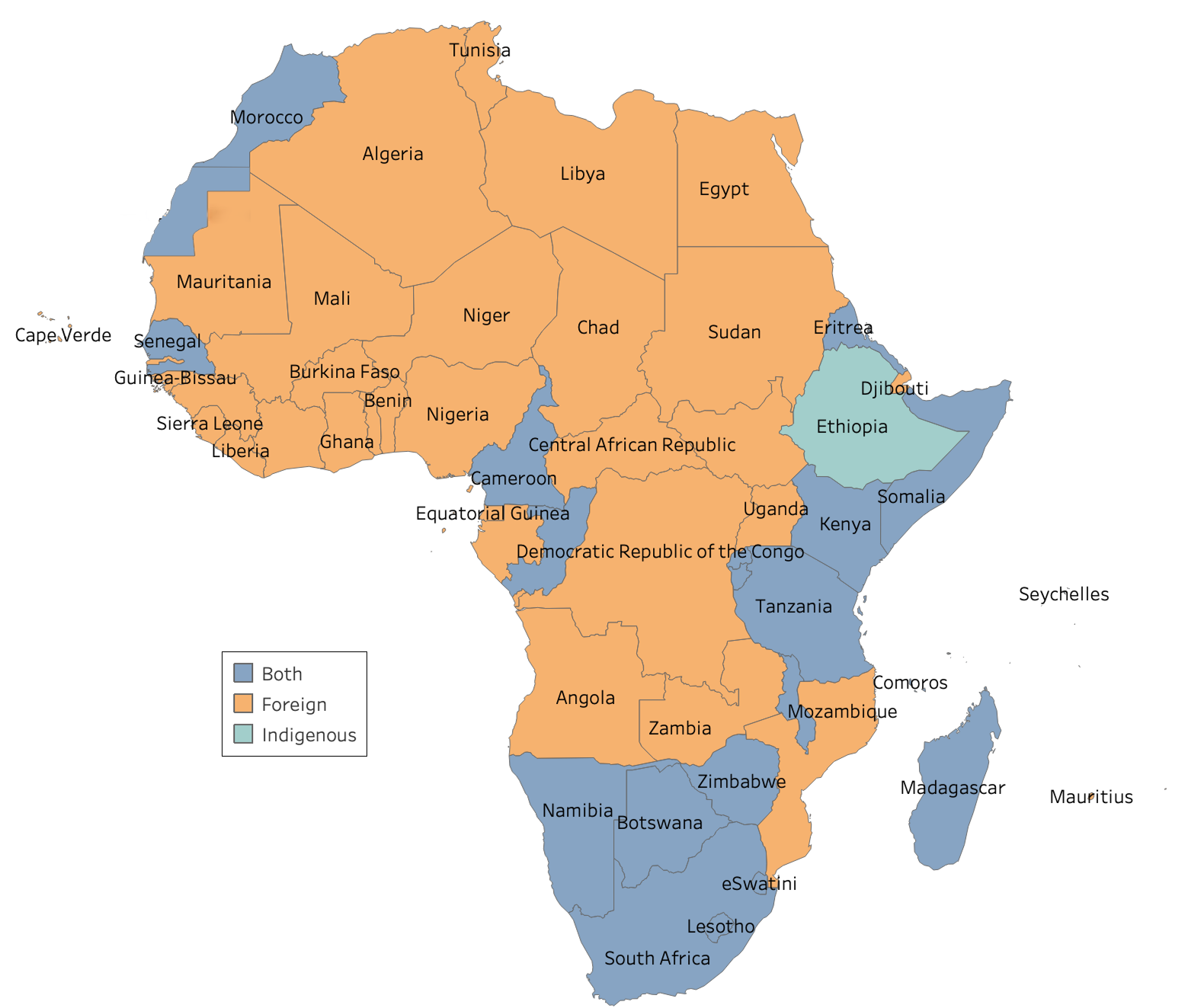}
    \justifying
    \small (b) Official national language policies across Africa.
    \label{fig:map_lang}
\end{minipage} \\[1em]
\begin{minipage}{0.53\textwidth}
    \centering
    \includegraphics[width=\textwidth]{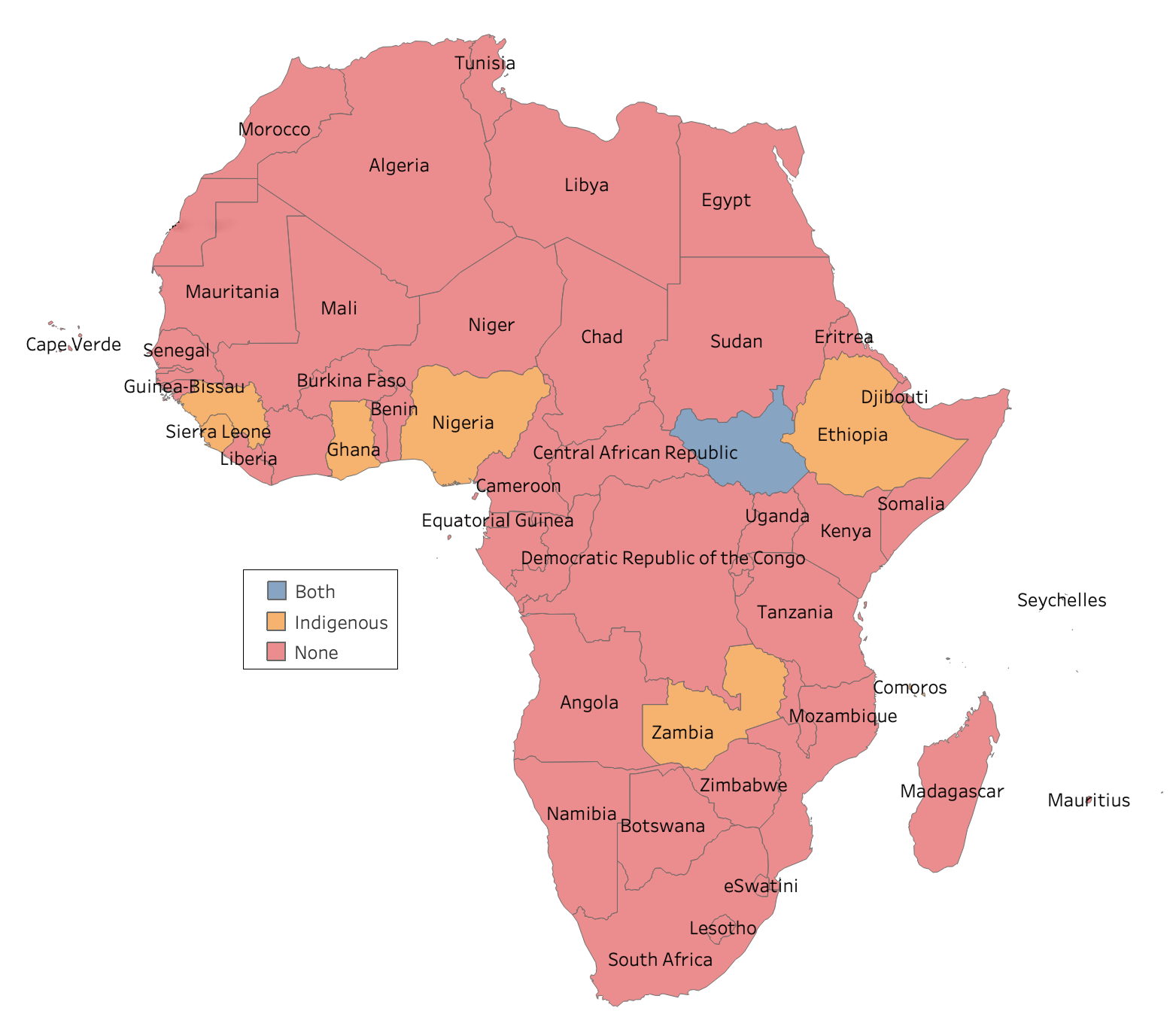}
    \justifying
    \small (c) Official regional language policies across Africa. \textit{None} means no additional policy is available on a regional level. 
    \label{fig:map_reg}
\end{minipage} &
\begin{minipage}{0.43\textwidth}
    \centering
    \includegraphics[width=\textwidth]{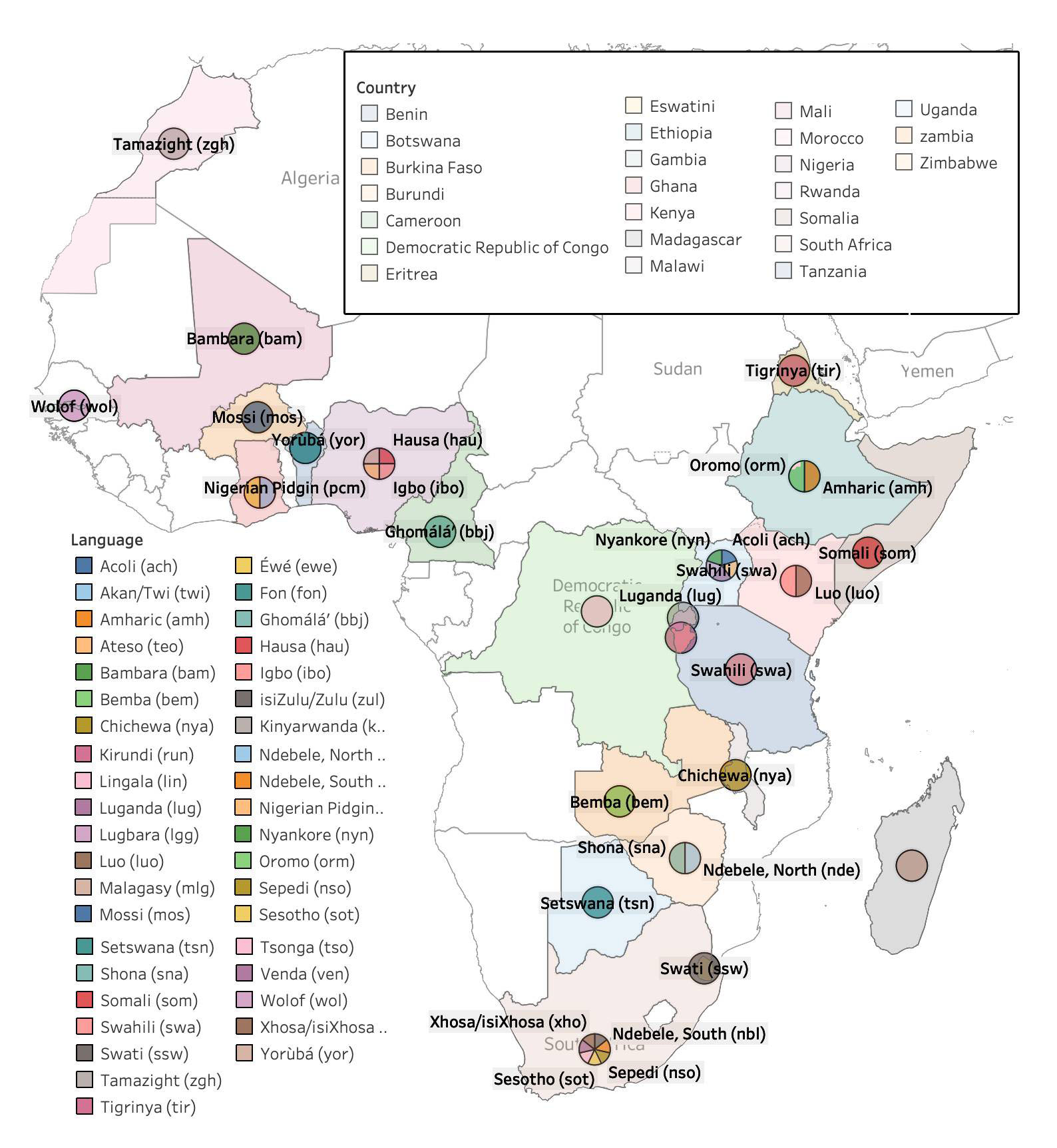}
    \justifying
    \small (d) We collect \ourbenchmark{} to empirically evaluate LLM performance on African languages, allowing us to demonstrate with evidence how current language policies directly impact progress in the field.

    \label{fig:map_data}
\end{minipage}
\end{tabular}%
}
\captionof{figure}{Maps of Africa showing the languages covered in this work and the language policies across the continent. The term \textit{Indigenous} language is broadly defined here as one that is `native' to the area~\cite{walsh2005will}. The term \textit{Both} in maps (a), (b), and (c) refers to \textit{Indigenous} and \textit{foreign} languages combined. Knowledge of what is an Indigenous language is based exclusively on \textit{Ethnologue} (\href{https://www.ethnologue.com}{https://www.ethnologue.com}).}
\label{fig:maps}
\end{strip}

\begin{abstract}
Africa’s rich linguistic heritage remains underrepresented in NLP, largely due to historical policies that favor foreign languages and create significant data inequities. In this paper, we integrate theoretical insights on Africa’s language landscape with an empirical evaluation using \ourbenchmark— a comprehensive benchmark we curate from large-scale, publicly accessible datasets capturing the continent’s linguistic diversity. By systematically assessing the performance of leading large language models (LLMs) on \ourbenchmark, we demonstrate how policy-induced data variations directly impact model effectiveness across African languages.\footnote{\ourbenchmark~is publicly available at \href{https://github.com/UBC-NLP/sahara}{https://github.com/UBC-NLP/sahara}.} Our findings reveal that while models perform reasonably well on few languages, many Indigenous languages remain marginalized due to sparse data. Leveraging these insights, we offer actionable recommendations for policy reforms and inclusive data practices. Overall, our work underscores the urgent need for a dual approach—combining theoretical understanding with empirical evaluation—to foster linguistic diversity in AI for African communities.
\end{abstract}

\section{Introduction}\label{sec:introduction}

The digital landscape of Africa is as diverse as its cultures and languages, yet the continent’s rich linguistic heritage remains largely underrepresented in NLP research. Despite significant transformations over the past decade driven by advances in large language models (LLMs) and a growing interest in linguistic diversity, most efforts have concentrated on a handful of widely spoken languages such as Swahili, Afrikaans, and Hausa, while many Indigenous languages continue to be sidelined due to the paucity of accessible, high-quality datasets. This oversight, deeply rooted in historical language policies that have neglected robust data collection and accessibility, has far-reaching implications for the development of language technology across Africa \cite{ethnologue}. Recent initiatives, including the  work of Masakhane \cite{dione-etal-2023-masakhapos, adelani-etal-2021-masakhaner, adelani-etal-2023-masakhanews, adelani2024irokobench, bandarkar-etal-2024-belebele} and the efforts of platforms like HuggingFace, have begun to address these challenges by curating language-specific datasets and adapting transfer learning techniques for low-resource African languages. Models such as Cheetah \cite{adebara-etal-2024-cheetah}, Serengeti \cite{adebara-etal-2023-serengeti}, Toucan \cite{elmadany-etal-2024-toucan}, Afri-XLMR \cite{alabi-etal-2022-adapting}, AfriTeVa \cite{oladipo-etal-2023-better, jude-ogundepo-etal-2022-afriteva}, mBERT \cite{DBLP:journals/corr/abs-1810-04805}, XLM-R \cite{conneau-etal-2020-unsupervised}, and LLaMA~\cite{touvron2023llama} have leveraged multilingual data to enhance performance on these languages, yet progress remains uneven—highlighting that the advancements achieved are primarily concentrated on a select few. 

In this study, we undertake a comprehensive empirical evaluation of leading LLMs on an extensive benchmark that we collect using mostly existing datasets to unravel how current African language policies, by dictating data availability and accessibility, directly shape model performance across the continent’s diverse linguistic spectrum. By systematically analyzing the distribution of existing datasets across languages and performance patterns among various LLMs, we not only pinpoint which languages benefit from ample data resources but also illuminate the underlying factors that contribute to their relative success. Our investigation reveals that the disparities in NLP outcomes are closely tied to policy-driven data inequities, offering concrete evidence of the need for more inclusive, forward-thinking language policies. These insights provide actionable recommendations for enhancing dataset creation and model training, ultimately aiming to bridge the digital divide and foster linguistic diversity in artificial intelligence for African communities.

Overall, we make the following contributions: \textbf{(1) Benchmark with wide, diverse coverage.} We introduce \textit{Sahara}, a comprehensive benchmark assembled from large-scale, publicly accessible, and inclusive datasets that capture the rich linguistic diversity of Africa. This benchmark enables a systematic evaluation of data availability and model performance across a broad range of African languages. \textbf{(2) Dynamic leaderboard.} We develop a dynamic leaderboard built on best design practices, providing a transparent and continually updated platform to track progress and benchmark innovations in African NLP. \textbf{(3) Empirical evaluation.} We perform an extensive empirical assessment of a wide spectrum of models thereby establishing robust baselines and revealing performance disparities linked to data resource availability. \noindent \textbf{(4) Policy-driven and actionable insights.} We uncover clear links between current language policies, data availability, and model performance, demonstrating how policy-induced data inequities shape the effectiveness of language models across Africa. Our analysis provides recommendations for future research and development by informing policy reforms that foster more inclusive language technologies tailored to the needs of African communities. We now discuss language polices in Africa.

\section{Language Policies in Africa}\label{sec:langpolicy}

Across Africa, the predominant response to the continent's multilingual landscape has been the adoption of a foreign language for official functions. However, in some instances, select Indigenous African languages receive official recognition at regional or national levels or within educational contexts~\cite{petzell2012linguistic, foster2021language, ouane2010and}. \footnote{An \textit{Indigenous} language is defined here as one that is \textit{native} to a particular region, rather than introduced from elsewhere, in contrast to a \textit{foreign} language, which may nonetheless serve as the mother tongue for certain local populations \cite{walsh2005will}. A \textit{national} language refers to a language spoken broadly across an entire country. A \textit{regional} language is spoken by a significant number of people within a specific area but does not have widespread national usage. An \textit{educational} language is officially designated as a medium of instruction within the educational system.} For instance in Nigeria, English is the official language, with only three out of $512$ Indigenous languages recognized at the regional level. Similarly, Ghana designates English as its official language while also recognizing ten of its $73$ Indigenous languages for institutional use. In Tanzania, Swahili is the sole Indigenous official language among $118$ languages, alongside English. Kenya grants official status to $12$ of its $61$ languages, while South Africa recognizes $12$ out of $20$ Indigenous languages as institutional languages~\cite{adebara-abdul-mageed-2022-towards}. Figures~\ref{fig:maps}(a), ~\ref{fig:maps}(b), ~\ref{fig:maps}(c), and ~\ref{fig:maps}(d) illustrate the distribution of language policies across the continent. 

Even when Indigenous languages are granted official status alongside foreign languages, they often serve a symbolic rather than functional role. For instance, although Kiswahili is recognized as an official language by the African Union, its website and official document releases remain in English and French. A similar pattern emerges in education systems: where Indigenous languages are used, their role is typically confined to early childhood education and is usually paired with a foreign language rather than serving as the sole medium of instruction~\cite{petzell2012linguistic, foster2021language, ouane2010and}. These language policies significantly shape language use across various domains, including newspapers, radio, television, and social media. Due to the dominance of foreign languages in official contexts, major newspapers and government publications are predominantly produced in languages such as English, French, and Portuguese, thereby limiting access for speakers of Indigenous languages~\cite{rescue_agbozo_2021}. Although some Indigenous languages are used on television and radio, foreign languages tend to dominate national broadcasts, especially in political discourse and formal news reporting~\cite{Cheo2023, myers2020local, nca2016list}.

Social media presents a more flexible environment where users can communicate in Indigenous languages; however, platform support remains uneven~\cite{sunday_etal_2018, molale_mpofu_2023, rescue_agbozo_2021}. Many African languages are underrepresented in digital spaces, lacking features such as text input options, spell checkers, or automated translation services. Consequently, users often resort to code-switching between Indigenous and dominant foreign languages, reinforcing existing linguistic divides. These trends underscore how official language policies shape broader language use, ultimately contributing to the marginalization of Indigenous languages in both public and digital communication spheres.

The analysis of language policies in Africa reveals how historical choices have led to significant data imbalances for Indigenous languages. Motivated by this, we built our benchmark,~\ourbenchmark~to evaluate how these policies directly impact data availability, coverage, and the performance of LLMs on African languages. In the following section, we detail the construction of ~\ourbenchmark~and demonstrate its role in linking language policy to empirical outcomes in NLP.

\section{Related Work}\label{sec:litreview}
The development of NLP for African languages is hindered by data scarcity, policy-induced disparities, and suboptimal model performance. Recent initiatives have sought to mitigate these challenges by curating diverse datasets, rigorously evaluating multilingual models, and advocating for more inclusive language policies. For example, several initiatives have contributed to expanding datasets for African \cite{yimametalcoling2020, muhammad-etal-2022-naijasenti, aliyu2022herdphobia, muhammadSemEval2023, muhammad-etal-2023-afrisenti, ilevbare2024ekohate, elmadany-etal-2024-toucan}. Some notable ones like \citet{adelani-etal-2021-masakhaner, adelani-etal-2022-masakhaner} introduced MasakhaNER, a named entity recognition dataset for 10 African languages, highlighting the limitations in existing resources. Similarly, IROKOBench \cite{adelani2024irokobench} evaluated LLMs on African languages, revealing stark performance gaps between high- and low-resource languages. \citet{adebara2022afrolid} developed AfroLID, a neural language identification model covering 517 African languages, showing that most languages lack NLP datasets beyond language identification.
\begin{table*}[!ht]
\centering
\resizebox{0.90\textwidth}{!}{%
\begin{tabular}{cllll}
\toprule
  \textbf{Cluster}   & \textbf{Task} & \textbf{Identifier} & \textbf{\#AL} & \textbf{\#DS} \\ \midrule

\multirow{5}{*}{\rotatebox[origin=c]{90}{Classification}}                            & Cross-Lingual Natural Language Inference~\cite{adelani2024irokobench}   &xnli                      & $16$                                       & $1$\\
                                                & Language Identification~\cite{adebara2022afrolid}               &langid               & $517$                                       & $1$\\ 
                                                 & News Classification~\cite{azime2021amharic, niyongabo-etal-2020-kinnews, davis_david_2020_4300294}  & news            & $4$                                       & $3$\\

                                                & Sentiment Analysis~\cite{diallo2021bambara, oyewusi2020semantic, shode2022yosm}   &sentiment             & $3$                                       & $3$\\ 
                                                & Topic Classification~\cite{hedderich-etal-2020-transfer}      &topic       & $2$                                       & $1$\\
                                                \midrule

 \multirow{4}{*}{\rotatebox[origin=c]{90}{Generation}}                 & 
                        Machine Translation~\cite{adelani-etal-2022-thousand, reid-etal-2021-afromt, ogueji2019pidginunmt, akera2022salt}                   &mt              & $29$                                       & $5$\\
                                                & Paraphrase~\cite{scherrer_yves_2020_3707949}     &paraphrase                      & $4$                                       & $1$\\

                                                & Summarization~\cite{hasan-etal-2021-xl, adebara2024cheetah}       & summary             & $10$                                       & $2$\\
                                                & Title Generation~\cite{hasan-etal-2021-xl, adebara2024cheetah}       & title-gen           & $10$                                       & $2$\\
 \midrule
 \multirow{4}{*}{\rotatebox[origin=c]{90}{MCCR}}  & General Knowledge~\cite{adelani2024irokobench}     & mmlu                   & $16$                                       & $1$\\ 

& Mathematical Word Problems~\cite{adelani2024irokobench}               & mgsm          & $16$                                       & $1$\\
                                                & Reading  Comprehension~\cite{bandarkar-etal-2024-belebele} & reading-comp& $25$ & $1$\\
                                                & Context-based Question Answering~\cite{clark-etal-2020-tydi}  & squad-qa                              & $1$                                       & $1$\\
                                                \midrule
\multirow{3}{*}{\rotatebox[origin=c]{90}{Tokens} } & NER~\cite{Adelani2021MasakhaNERNE, Adelani2022MasakhaNER2A, eiselen-2016-government, alabi-etal-2020-massive, pan-etal-2017-cross}    & ner                          & $27$                                       & $5$\\
                                                & Phrase Chunking~\cite{eiselen-2016-government}  &chunking                 & $8$                                       & $1$\\
                                                & POS Tagging~\cite{10.1145/3146387, 10.1145/3314942}  &pos                     & $1$                                       & $1$\\
                                                \midrule

\multicolumn{1}{c}{\textbf{Total}}                                             & \multicolumn{1}{l}{$\bf 16$}            &                                       & $\bf 517$   & $\bf 30$                                    \\ \bottomrule                            
\end{tabular}%
}

\caption{Descriptive statistics of languages and dataset diversity in \ourbenchmark~across different tasks and task-clusters. \textbf{\#AL:} African Languages, \textbf{\#DS:} datasets, \textbf{MCCR:} multiple-choice, comprehensive and reasoning task-cluster.}
\label{tab:benchmark_tasks}
\end{table*}

Language policies play a crucial role in determining resource availability.~\citet{adebara-abdul-mageed-2022-towards, hershcovich-etal-2022-challenges} emphasized the sociolinguistic factors affecting NLP development, noting that languages with official status receive more institutional support, fostering greater digital representation.~\citet{bird-2020-decolonising} further critiqued the Western-centric approach in speech and language technologies, calling for AI models that align with Indigenous linguistic contexts. Model performance in African languages remains uneven, largely due to the dominance of high-resource languages in multilingual training. \citet{conneau2018xnli} introduced XLM-R, demonstrating the limitations of cross-lingual transfer learning for underrepresented languages. Meta AI’s No Language Left Behind (NLLB) project  \cite{nllb2022} attempted to improve multilingual models for low-resource languages, but performance remained inconsistent, particularly in generation tasks. Despite these advances, African NLP still requires increased investment in dataset curation, inclusive policy frameworks, and model development tailored to the linguistic diversity of the continent. We now introduce~\ourbenchmark.

\section{\includegraphics[scale=0.04]{Images/sahara_logo.png} Sahara Benchmark}\label{sec:benchmark}

Our objective is to create a comprehensive benchmark for African NLP that enables (i) analysis of existing language resources and (ii) the assessment of language models and tracking of the progress of African NLP. To achieve this goal, we develop \ourbenchmark, adhering to several key design principles that we will now elucidate. \ourbenchmark~establishes a comprehensive and adaptable benchmark for African NLP using publicly available datasets. Figure~\ref{fig:map_data} shows coverage of~\ourbenchmark.


\paragraph{Wide and Diverse Coverage.}
\ourbenchmark~is the most comprehensive benchmark for African NLP, covering a vast range of languages. We achieve this by collecting high-quality, publicly available datasets from as many African languages as possible, ensuring easy accessibility for researchers evaluating their models. As a result, \ourbenchmark~supports $517$ languages across various tasks, making it the most extensive and representative benchmark for African NLP. It encompasses languages from $50$ out of $54$ African countries and includes data written in five different scripts: \textit{Arabic, Coptic, Ethiopic, Latin}, and \textit{Vai}, spanning five language families across the continent.

\paragraph{Tasks and Task Clusters.}  
\ourbenchmark~is designed to support a broad spectrum of NLP tasks, which we systematically organize into coherent task clusters. The inclusion of well-defined, challenging clusters enables a more meaningful assessment of a model’s capabilities and allows researchers to evaluate performance on specific clusters. As shown in Table~\ref{tab:benchmark_tasks}, \ourbenchmark~comprises four clusters: \textit{text classification}, \textit{text generation}, \textit{multiple-choice, comprehensive and reasoning (MCCR)}, and \textit{token-level classification}. For further details on the datasets used for each task, please refer to Appendix \ref{appendix: task_cluster_detail}.

\begin{table*}[]
\resizebox{\textwidth}{!}{%
\begin{tabular}{cllcrrrrrrlrrrrrrrrrrrrrrrrlrr}
\toprule
\multicolumn{1}{c}{}                                                & \multicolumn{1}{c}{}                                      & \multicolumn{1}{c}{}                                  & \multicolumn{1}{c}{}                                  & \multicolumn{6}{c}{\textbf{SLM}}                                                                                                                                                                          & \textbf{} & \multicolumn{16}{c}{\textbf{LLM}}                                                                                                                                                                                                                                                                                                                                                                                                                                                                                                                                                                                                                                                                                                                                                                                                               & \textbf{}                                                & \multicolumn{2}{c}{\textbf{Closed LLM}}                                                                                       \\ \cmidrule{5-10} \cmidrule{12-27} \cmidrule{29-30}
\multicolumn{1}{c}{\multirow{-2}{*}{\textbf{Cluster}}}              & \multicolumn{1}{c}{\multirow{-2}{*}{\textbf{Identifier}}} & \multicolumn{1}{c}{\multirow{-2}{*}{\textbf{Metric}}} & \multicolumn{1}{c}{\multirow{-2}{*}{\textbf{Shots}}} & \multicolumn{1}{c}{\textbf{\colorbox{yellow!20}{\textbf{S$_1$}}}} & \multicolumn{1}{c}{\textbf{\colorbox{yellow!20}{\textbf{S$_2$}}}} & \multicolumn{1}{c}{\textbf{\colorbox{yellow!20}{\textbf{S$_3$}}}} & \multicolumn{1}{c}{\textbf{\colorbox{yellow!20}{\textbf{S$_4$}}}} & \multicolumn{1}{c}{\textbf{\colorbox{yellow!20}{\textbf{S$_5$}}}} & \multicolumn{1}{c}{\textbf{\colorbox{yellow!20}{\textbf{S$_6$}}}} & \textbf{} & \multicolumn{1}{c}{\textbf{\colorbox{blue!20}{\textbf{L$_1$}}}} & \multicolumn{1}{c}{\textbf{\colorbox{blue!20}{\textbf{L$_2$}}}} & \multicolumn{1}{c}{\textbf{\colorbox{blue!20}{\textbf{L$_3$}}}} & \multicolumn{1}{c}{\textbf{\colorbox{blue!20}{\textbf{L$_4$}}}}   & \multicolumn{1}{c}{\textbf{\colorbox{blue!20}{\textbf{L$_5$}}}}   & \multicolumn{1}{c}{\textbf{\colorbox{blue!20}{\textbf{L$_6$}}}}                               & \multicolumn{1}{c}{\textbf{\colorbox{blue!20}{\textbf{L$_7$}}}}                               & \multicolumn{1}{c}{\textbf{\colorbox{blue!20}{\textbf{L$_8$}}}}                               & \multicolumn{1}{c}{\textbf{\colorbox{blue!20}{\textbf{L$_9$}}}} & \multicolumn{1}{c}{\textbf{\colorbox{blue!20}{\textbf{L$_{10}$}}}}                              & \multicolumn{1}{c}{\textbf{\colorbox{blue!20}{\textbf{L$_{11}$}}}}                              & \multicolumn{1}{c}{\textbf{\colorbox{blue!20}{\textbf{L$_{12}$}}}}                              & \multicolumn{1}{c}{\textbf{\colorbox{blue!20}{\textbf{L$_{13}$}}}} & \multicolumn{1}{c}{\textbf{\colorbox{blue!20}{\textbf{L$_{14}$}}}}                              & \multicolumn{1}{c}{\textbf{\colorbox{blue!20}{\textbf{L$_{15}$}}}}                              & \multicolumn{1}{c}{\textbf{\colorbox{blue!20}{\textbf{L$_{16}$}}}}                              & \textbf{}                                                & \multicolumn{1}{c}{\textbf{\colorbox{red!20}{\textbf{CL$_1$}}}}                              & \multicolumn{1}{c}{\textbf{\colorbox{red!20}{\textbf{CL$_2$}}}}                             \\ \midrule
 \multirow{6}{*}{\rotatebox[origin=c]{90}{\textbf{Classification}}} & xlni & Acc. & 5 & 33.6 & 33.7 & 36.7 & 41.18 & 40.4 & 46.68 &  & 48.08 & 55.06 & 34.3 & 40.69 & 37.09 & 61.09 & 57.86 & \underline{65.58} & 44.2 & 38.19 & 51.89 & 48.38 & 52.09 & 35.87 & 46.77 & 59.67 &  & 69 & \colorbox{green!20}{\textbf{69.6}}\\
 & lid & $F_1$ &10& 0 & 0 & 0.87 & 0.26 & 0.85 & 0.77 &  & 2.82 & 3.6 & 0 & 1.49 & 2.27 & 1.95 & \colorbox{green!20}{\underline{\textbf{4.61}}} & 4 & 1.81 & 3.34 & 3.26 & 3.07 & 3.3 & 3.05 & 3.32 & 4.57 &  & 3.73 & 4.48\\
 & news & $F_1$ & 3 & 0.86 & 0.35 & 0.15 & 17.54 & 0 & 6.64 &  & 21.94 & 0 & 2.06 & 24.06 & 0.12 & 25.83 & 7.83 & \underline{35.56} & 28.03 & 0.61 & 35.26 & 33.31 & 34.81 & 3.45 & 30.46 & 28.66 &  & \colorbox{green!20}{\textbf{40.09}} & 32.19\\
 & sentiment & $F_1$ & 5 & 38.51 & 24.75 & 30.45 & 11.05 & 28.77 & 1.99 &  & 36.09 & 42.47 & 38.4 & 28.36 & 36.07 & 34.56 & 20.55 & 46.35 & 35.86 & 25.94 & 24.51 & 45.82 & 35.89 & 19.98 & 37.07 & \underline{47.14} &  & 55.67 & \colorbox{green!20}{\textbf{57.52}}\\
 & topic & $F_1$ & 3 & 8.73 & 31.62 & 20.3 & 19.79 & 12.46 & 17.15 &  & 48.11 & 39.35 & 34.95 & 28.55 & 42.12 & 44.3 & 31.8 & \underline{70.73} & 36.47 & 43.74 & 64.89 & 56.48 & 48.73 & 52.09 & 50.96 & 53.16 &  & 67.93 & \colorbox{green!20}{\textbf{76.57}}\\ \cmidrule{3-30}
 &  & \textbf{Avg.} &  & 16.34 & 18.08 & 17.69 & 17.96 & 16.50 & 14.65 &  & 31.41 & 28.10 & 21.94 & 24.63 & 23.53 & 33.55 & 24.53 & \underline{44.44} & 29.27 & 22.36 & 35.96 & 37.41 & 34.96 & 22.89 & 33.72 & 38.64 &  & 47.28 & \colorbox{orange!20}{\textbf{48.07}}\\ \midrule
\multirow{7}{*}{\rotatebox[origin=c]{90}{\textbf{Generation}}}  & mt\_eng2xx & spBleu\textsuperscript{1K} & 5 & 1.79 & 3.63 & 3.45 & 3.25 & 2.12 & 3.4 &  & 7.07 & 7.61 & 1.72 & 4.37 & 2.79 & 6.24 & 6.91 & 4.56 & 3.73 & 3.29 & 8.84 & 8.7 & \underline{9.36} & 1.4 & 6.06 & 8.22 &  & 12.18 & \colorbox{green!20}{\textbf{12.36}}\\
 & mt\_fra2xx & spBleu\textsuperscript{1K} & 5 & 0.79 & 0.36 & 0.74 & 1.05 & 1.25 & 1.32 &  & 0.73 & 0.95 & 0.7 & 1.19 & 0.66 & 2.03 & 1.67 & 2.05 & 1.15 & 0.82 & 1.93 & 1.87 & 1.77 & 1.05 & 1.04 & \underline{2.3} &  & 3.41 & \colorbox{green!20}{\textbf{3.42}}\\
 & mt\_xx2xx & spBleu\textsuperscript{1K} & 5 & 0.4 & 0.45 & 0.42 & 0.42 & 0.36 & 0.18 &  & 0.43 & 0.63 & 0.72 & 0.55 & 0.46 & 0.71 & 0.9 & \underline{1.39} & 0.55 & 0.17 & 1.28 & 1.05 & 1.37 & 0.67 & 0.38 & 0.85 &  & \colorbox{green!20}{\textbf{5.15}} & 4.44\\
 & paraphrase & spBleu\textsuperscript{1K} & 5 & 23.2 & 28.88 & 31.15 & 19.12 & 25.93 & 17.85 &  & 28.73 & 25.83 & 31.68 & 31.33 & 23.66 & 17.26 & 21.34 & 15.37 & 32.13 & 30.68 & 26.86 & 19.64 & 27.79 & 19.23 & \colorbox{green!20}{\underline{\textbf{36.06}}} & 26.06 &  & 23.2 & 20.35\\
 & summary & rougeL & 2 & 10.03 & 3.24 & 4.37 & 10.07 & 0.91 & 13.34 &  & 15.1 & 10.43 & 1.45 & 12.74 & 5.92 & 16.16 & 15.73 & 17.13 & 13.27 & 4.19 & 16.92 & \colorbox{green!20}{\underline{\textbf{17.19}}} & 16.4 & 15.46 & 14.88 & 16.09 &  & 5.99 & 16.3\\
 & title & spBleu\textsuperscript{1K} & 2 & 2.14 & 2.81 & 0.25 & 2.3 & 0.03 & 5.78 &  & 7.48 & 5.78 & 3 & 5.94 & 1.58 & 7.75 & 7.51 & 8.65 & 6.83 & 3.2 & \underline{11.07} & 10.34 & 8.78 & 7.32 & 6.69 & 8.65 &  & \colorbox{green!20}{\textbf{13.62}} & 9.48\\ \cmidrule{3-30}
 &  & \textbf{Avg.} &  & 6.39 & 6.56 & 6.73 & 6.04 & 5.10 & 6.98 &  & 9.92 & 8.54 & 6.55 & 9.35 & 5.85 & 8.36 & 9.01 & 8.19 & 9.61 & 7.06 & \colorbox{orange!20}{\underline{\textbf{11.15}}} & 9.80 & 10.91 & 7.52 & 10.85 & 10.36 &  & 10.59 & 11.06\\  \midrule
\multirow{5}{*}{\rotatebox[origin=c]{90}{\textbf{MCCR}}} & mmlu & Acc. & 5 & 24.09 & 28.5 & 28.68 & 27.29 & 28.9 & 30.99 &  & 31.88 & 40.9 & 26.4 & 44.51 & 22.1 & 42.09 & 44.9 & 60.71 & 51.5 & 24 & 56.51 & 59.3 & 50.42 & 59.81 & 29.8 & \colorbox{green!20}{\underline{\textbf{61.91}}} &  & 81.6 & 58.6\\
 & mgsm & ExactM & 5 & 2.6 & 3.21 & 0 & 1.2 & 3.39 & 0.7 &  & 2.9 & 2.8 & 3.3 & 2.7 & 1.5 & 2.3 & 5.3 & 0.6 & 4.4 & 3 & 10.1 & 7.51 & 7.6 & 4.5 & 4.39 & \underline{11.1} &  & \colorbox{green!20}{\textbf{45.9}} & 29.5\\
 & belebele & Acc. & 5 & 27.71 & 14.71 & 25.2 & 29.31 & 31.31 & 24.69 &  & 22.7 & 26.93 & 25.09 & 28.24 & 26.33 & 25.63 & 31.04 & 32.34 & 33.64 & 26.61 & 30.4 & \underline{35.92} & 33.72 & 30.83 & 20.99 & 29.12 &  & 36.6 & \colorbox{green!20}{\textbf{39.7}}\\
 & squad\_qa & $F_1$ & 5 & 57.45 & 70.89 & 67.32 & 43.13 & 71.31 & 74.97 &  & 78.04 & 80.91 & 52.66 & 59.48 & 63.41 & 73.78 & \colorbox{green!20}{\underline{\textbf{82.49}}} & 79.16 & 69.25 & 64.64 & 77.65 & 76.35 & 66.25 & 53.92 & 74.64 & 80.05 &  & 78.02 & 76.11\\ \cmidrule{3-30}
 &  & \textbf{Avg.} &  & 27.96 & 29.33 & 30.30 & 25.23 & 33.73 & 32.84 &  & 33.88 & 37.89 & 26.86 & 33.73 & 28.34 & 35.95 & 40.93 & 43.20 & 39.70 & 29.56 & 43.67 & 44.77 & 39.50 & 37.27 & 32.46 & \underline{45.55} &  & \colorbox{orange!20}{\textbf{60.53}} & 50.98\\  \midrule
\multirow{4}{*}{\rotatebox[origin=c]{90}{\textbf{Tokens}}} & ner & $F_1$ & 5 & 0 & 0.78 & 1.76 & 2.99 & 0.13 & 3.65 &  & 2.26 & 7.49 & 2.35 & 3.96 & 0.2 & 6.03 & 13.14 & 11.4 & 8.51 & 1.52 & 6.86 & 10.54 & 8.64 & 6.04 & 4.3 & \underline{16.88} &  & \colorbox{green!20}{\textbf{35.06}} & 27.04\\
 & phrase & $F_1$ & 5 & 14.79 & 28.5 & 29.44 & 11.52 & 24.91 & 27.73 &  & 15.95 & 29.35 & 29.42 & 16.55 & 30.87 & 13.59 & 30.01 & 23.97 & 26.34 & 23.09 & 25.55 & 23.13 & 16.78 & 25.13 & 29.44 & \underline{31.05} &  & \colorbox{green!20}{\textbf{36.69}} & 36.45\\
 & pos & $F_1$ & 5 & 8.83 & 6.73 & 10.56 & 8.04 & 10.3 & 10.46 &  & 9.13 & 13.2 & 9.81 & 11.68 & 12.52 & 13.86 & 19 & 13.99 & 18.36 & 11.4 & 14.13 & 15.04 & 11.52 & 20.76 & 13.71 & \underline{27.54} &  & \colorbox{green!20}{\textbf{62.84}} & 38.67\\ \cmidrule{3-30}
 &  & \textbf{Avg.} &  & 7.87 & 12.00 & 13.92 & 7.52 & 11.78 & 13.95 &  & 9.11 & 16.68 & 13.86 & 10.73 & 14.53 & 11.16 & 20.72 & 16.45 & 17.74 & 12.00 & 15.51 & 16.24 & 12.31 & 17.31 & 15.82 & \underline{25.16} &  & \colorbox{orange!20}{\textbf{44.86}} & 34.05\\ \midrule
 &  & \textbf{Overall} &  & 14.64 & 16.49 & 17.16 & 14.19 & 16.78 & 17.10 &  & 21.08 & 22.80 & 17.30 & 19.61 & 18.06 & 22.25 & 23.80 & 28.07 & 24.08 & 17.75 & 26.57 & 27.05 & 24.42 & 21.25 & 23.21 & \underline{29.93} &  & \colorbox{orange!20}{\textbf{40.82}} & 36.04\\ \bottomrule
\end{tabular}%
}
\caption{Few-shot evaluation across different task clusters/tasks. The highest score for each individual task is in \colorbox{green!20}{\textbf{bold green}}, while the best average score across each task cluster (and \textbf{overall} score, calculated as average of task cluster averages) is in \colorbox{orange!20}{\textbf{bold orange}}. \underline{\textbf{Underline}}: refers to the best score across the open source/weights LLMs. \textbf{Shots}: number of shots. \textbf{SLMs:} \colorbox{yellow!20}{\textbf{S$_1$}} Llama3.2 (1B), \colorbox{yellow!20}{\textbf{S$_2$}} Llama3.2 (3B), \colorbox{yellow!20}{\textbf{S$_3$}} Gemma2 (2B), \colorbox{yellow!20}{\textbf{S$_4$}} Phi-3.5 (3.8B), \colorbox{yellow!20}{\textbf{S$_5$}} Phi-4 (3.8B), and \colorbox{yellow!20}{\textbf{S$_6$}} Gemma3 (4B). \textbf{LLMs:} \colorbox{blue!20}{\textbf{L$_1$}} Llama3.1 (8B), \colorbox{blue!20}{\textbf{L$_2$}} Gemma2 (9B), \colorbox{blue!20}{\textbf{L$_3$}} Aya (8B), \colorbox{blue!20}{\textbf{L$_4$}} Babel (9B), \colorbox{blue!20}{\textbf{L$_5$}} Command-R-Plus-R7B (8B), \colorbox{blue!20}{\textbf{L$_6$}} Gemma3 (12B), \colorbox{blue!20}{\textbf{L$_7$}} Gemma2 (27B), \colorbox{blue!20}{\textbf{L$_8$}} Gemma3 (27B), \colorbox{blue!20}{\textbf{L$_9$}} DeepSeek-R1-Distill-Qwen (32B), \colorbox{blue!20}{\textbf{L$_{10}$}} Aya (35B), \colorbox{blue!20}{\textbf{L$_{11}$}} Llama3.1 (70B), \colorbox{blue!20}{\textbf{L$_{12}$}} Llama3.3 (70B), \colorbox{blue!20}{\textbf{L$_{13}$}} DeepSeek-R1-Distill-Qwen (70B), \colorbox{blue!20}{\textbf{L$_{14}$}} Babel (83B), \colorbox{blue!20}{\textbf{L$_{15}$}} Command-R-Plus (104B), \colorbox{blue!20}{\textbf{L$_{16}$}} Command-A (111B). \textbf{CLosed LLMs:} \colorbox{red!20}{\textbf{CL$_1$}} Claude-4-Sonnet, \colorbox{red!20}{\textbf{CL$_2$}} GPT-4.1. Table~\ref{appd_tab:results-subset} is a full-page version of this table.}
\label{tab:results-subset}
\end{table*}

\paragraph{Modular Public Leaderboard.}  
We implement a \textit{user-friendly leaderboard} for model evaluation on \ourbenchmark, hosted on Hugging Face (HF) Spaces. Evaluated models must be publicly available in HF repositories to ensure seamless integration. When submitting a model, users are required to provide detailed \textit{metadata}—including whether the model is pretrained, further pretrained, or instructed—and then submit an evaluation request. The submission enters a processing queue, and once evaluation is complete, the results are automatically displayed on the leaderboard. Our team will actively maintain the leaderboard by continuously integrating new publicly available tasks and datasets to keep it \textit{up-to-date}. \ourbenchmark\ thus serves as a valuable tool for assessing model performance in African NLP, accelerating the development of high-performance models, and contributing to the field’s growth. In addition to evaluating models on the entire benchmark, users have the flexibility to assess performance on specific task clusters, encouraging the development of specialized models tailored to address the unique challenges of African NLP.

\section{Experiments and Results}\label{sec:evaluation}
\subsection{Experimental Setup}

To comprehensively assess the progress of African NLP, we systematically evaluate a range of widely used instruction-based LLMs in a few-shot setting on the \ourbenchmark{} benchmark. We include both open and closed LLMs, such as Phi-3.5 and Phi-4~\cite{abdin2024phi3technicalreporthighly}, Gemma family~\cite{gemmateam2024gemma2improvingopen} models including Gemma-2 (2, 9, 27B) and Gemma-3 (4, 12, 27B), Llama family models~\cite{dubey2024llama} including Llama-3.1 (8, 70B), Llama-3.2 (1, 3B), and Llama-3.3 (70B), Aya models (8, 23B)~\cite{aryabumi2024aya23openweight}, and Command family models~\cite{cohere2025command} including Command-R-Plus-R7B, Command-R-Plus (104B) and Command-A (111B), as well as proprietary models such as GPT-4.1\footnote{\textbf{GPT-41.1 version:} gpt-4.1-2025-04-14.}
~\cite{achiam2023gpt} and Claude-4-Sonnet\footnote{\textbf{Claude-4-Sonnet version:} claude-4-sonnet-20250514}
~\cite{anthropic2024claude3}. We further analyze performance by categorizing models according to \textit{size}: small language models (\texttt{SLM}, $<$8B parameters) and large language models (\texttt{LLM}, $\geq$8B parameters), allowing investigation of the relationship between scale and effectiveness. For robust comparison, we aggregate evaluation datasets according to task clusters described in Appendix~\ref{appendix: task_cluster_detail}. From each task, we randomly sample $1,000$ data points for few-shot testing, employing well-curated, concise prompts—examples of which appear in Appendix~\ref{fig:few-shot}.

\subsection{Results and Discussion}

We evaluate $24$ LMs across four task clusters, maintaining a consistent assessment approach while varying evaluation metrics. Depending on the task, we employ \textit{Exact Match}, \textit{F\textsubscript{1}}, \textit{Accuracy}, \textit{spBLEU\textsuperscript{1K}}~\cite{elmadany-etal-2024-toucan}, and \textit{RougeL} scores.

As shown in Table~\ref{tab:results-subset}, in the \textit{classification tasks}, the closed models Claude-4-Sonnet and GPT-4.1 consistently lead in performance, followed by larger LLMs such as Gemma3 (27B) and CommandR-Plus-104B. In the \textit{language identification task}, most models struggle to grasp the task effectively, with only the larger LLMs demonstrating some understanding, suggesting that model size is crucial for achieving higher accuracy across various multilingual tasks. In the \textit{generation task cluster}, larger models perform as expected, with GPT-4.1 and Llama-3.1 (70B) achieving the best results. There is a noticeable performance increase across all models in the paraphrase task, while the summarization task exhibits a more balanced performance among all models. Considering their size, Phi-3.5 and Gemma3 (4B) perform well within this task cluster. In the \textit{MCCR tasks}, Command-A (111B) performs competitively against the closed Claude-4-Sonnet model. In the MGSM task, which involves solving mathematical word problems, most open-source models perform poorly, while Claude-4-Sonnet excels, achieving an exact match score of $22.34$\%. In token-level tasks, GPT-4.1 and Command-A (111B) consistently deliver strong performances, whereas the Aya model family struggle to grasp the tasks. 

Results demonstrate a nuanced landscape of performance across multilingual and African LLMs, with notable variations depending on parameter size and task type. In few-shot settings, the closed model Claude-4-Sonnet achieves the highest overall average score (40.82), outperforming all open models across most task clusters. Among open models, Command-A (111B) and Gemma-3 (27B) also exhibit strong performance, particularly in downstream tasks such as reasoning tasks, where Gemma2 (27B) reaches top score of 82.49 on Squad-QA and Command-A achieves 61.91 on MMLU task. For smaller parameter regimes, Gemma-2 (2B) and Phi-4 (3.8B) demonstrate the best performance on their model size. Furthermore, Gemma2 (9B, 12B) show consistent mid-to-high performance across various clusters, but do not surpass the larger models in aggregate. Importantly, translating \textit{into} English is consistently easier for the models than translating \textit{from} French, likely reflecting disparities in training data distribution and language representation on the target side. These findings highlight not only the dominance of closed LLMs and very large open models in multilingual tasks, but also the critical influence of resource availability and model scale on task-specific outcomes.

\paragraph{Which LM should I use for reasoning tasks?} In a limited resource setting, for mathematical QA tasks that fall in the \texttt{MCCR} cluster, Phi-4 ranks top among SLMs for 16 African Languages. In the LLMs category, Command-A (111B) has better performance on a higher number of African languages. For reading comprehension tasks (belebele), Phi-4 is most suitable in constrained compute resource settings while in scenarios without limits in compute resources, Llama 3.3-70B gives better performance.

\paragraph{Which African language(s) do LMs perform best on?}

The models achieve their highest performance on a small set of well-resourced African languages—chiefly Hausa, Swahili, Yor\`{u}b\'{a}, and Afrikaans—across diverse NLP tasks. Both open-source models like Command-A (111B) and Aya-35B and proprietary models such as Claude-4-Sonnet consistently excel on these languages, reflecting their ample labeled and parallel data. This trend spans simple classification and NER tasks to complex reasoning benchmarks (e.g., MMLU, XLNI) and machine translation, underscoring that data availability, rather than linguistic typology, is the primary driver of model effectiveness.

Models explicitly designed with African linguistic diversity, such as Aya-35B, also achieve strong results on Southern Sotho and Lingala—languages that have benefited from recent, community-led data initiatives. Swahili, in particular, ranks among the top languages for Llama3 models, likely due to its standardized orthography, regular morphology, and extensive bilingual corpora. Afrikaans and Swahili similarly perform well within Gemma2 and Command-R-Plus (104B), supported by their inclusion in massive multilingual datasets (e.g., mC4, OSCAR, WMT) and, in Afrikaans’s case, established government-supported digital media.

Overall, our findings reveal that LM performance correlates strongly with the volume and diversity of available data. Underperformance on low-resource African languages is not a function of inherent complexity but of underrepresentation in training data. Bridging this gap will require concerted efforts to collect, annotate, and standardize datasets that capture the dialectal, sociolinguistic, and orthographic richness of all African languages—an essential step toward improved model equity and real-world utility across the continent's full linguistic spectrum.

\section{Data-Performance-Policy Analysis}~\label{sec:analysis}
 \noindent To understand the relationship between policy, data availability, and performance, we carry out a number of analyses inspired by~\ourbenchmark~ data and empirical results on it. 

\paragraph{Data Availability and Diversity.}
As shown in Figure \ref{fig:cluster_lang}, among the $517$ languages in our benchmark, only $45$ have more than one dataset, while the majority are represented solely by language identification data. Amharic leads with $11$ unique datasets across all clusters, followed by Yor\`{u}b\'{a} with ten and Hausa with nine. A total of six languages have only one dataset beyond language identification, while the remaining $36$ have between two and eight datasets each. \textit{It is important to note that the disparity in dataset availability is closely linked to language policy.} The $45$ languages with multiple datasets generally hold official status or are spoken by a significant majority in their respective countries. Government policies that designate certain languages as official often yield greater institutional support, driving increased documentation, education, and media presence. For example, Amharic’s official status in Ethiopia contributes to its relatively rich dataset availability, while Yor\`{u}b\'{a} and Hausa, both widely used in Nigeria for regional governance and media, similarly benefit from enhanced linguistic resources. Conversely, languages lacking official recognition or widespread institutional support receive minimal investment in digital and linguistic infrastructure. The dominance of a few languages within policy frameworks perpetuates data scarcity for many Indigenous African languages, thereby limiting their inclusion in NLP advancements and reinforcing digital inequalities. 

It is noteworthy that availability of NLP data is not necessarily determined by the number of speakers and that other variables including language prestige, literacy rates, and economic incentives can be involved. This becomes evident when comparing speaker populations with actual data availability. For example, Hausa (73 million speakers), Swahili (80 million), and Zulu (28 million) are categorized as \textit{hopefuls—languages} with some available labeled datasets. In contrast, Naija (Nigerian Pidgin)—despite having approximately 153 million speakers—is classified as a \textit{left-behind} language~\cite{joshi-etal-2020-state}, largely due to decades of stigmatization and negative public attitudes~\cite{oyebola2023attitudes}, which have significantly contributed to the scarcity of accessible and annotated linguistic resources. Notably, several non-African languages with significantly smaller speaker populations—such as Catalan (5 million), Finnish (10 million), and Swedish (10.5 million)—are regarded as high-resource, primarily as a result of sustained efforts in documentation, digitization, and technological integration. This discrepancy underscores a broader systemic issue in language representation: \textit{it is not just linguistic demographics that shape digital inclusion, but also factors such as historical language prestige, policy decisions, digital infrastructure, and the extent of prior documentation}. 

These intersecting factors ultimately determine whether a language is integrated into AI systems or remains marginalized. Addressing this imbalance requires a shift in policy and practice to prioritize equitable digital representation. This includes targeted investments in multilingual data collection, increased funding for research on historically excluded and low-resource languages, and the intentional integration of these languages into education systems, media platforms, and digital technologies. Such efforts must also consider the sociopolitical and infrastructural barriers that shape language inclusion in AI, ensuring that underrepresented languages are not only documented but actively supported within the broader digital ecosystem.


\paragraph{Data Quality.}
Most of the downstream task datasets  are limited to simple tasks that require merely an atomic class label at the word, sentence, or document level (e.g., sentiment analysis, named entity recognition, topic classification). Furthermore, many of these datasets are translations of existing resources or content from high-resource languages like English and French, and they do not always reflect authentic language usage within the respective communities. For instance, AfriXLNI, AfriMMLU, and AfriMGSM \cite{adelani2024irokobench} were created by translating XLNI, MMLU, and MGSM into African languages. In these datasets, the frequent use of borrowed words and phonologized variants of foreign terms \cite{adebara-abdul-mageed-2022-towards} highlights the challenges posed by cross-lingual concept gaps. When translations are applied to classification tasks, misalignments in the attached labels can occur because labels are not always equivalent across languages. Some labels may not exist in the target language or may have restricted usage. For example, while English possesses a productive set of adjectives, languages like Yor\`{u}b\'{a} have a more limited adjective inventory that is context-dependent, meaning that an English adjective may be translated as a noun or verb in Yor\`{u}b\'{a} rather than as an adjective. 

\begin{figure*}[!ht]
    \centering
    \resizebox{0.974\textwidth}{!}{ 
        \begin{minipage}{\linewidth} 
            \centering
            \subfigure[Generation tasks.]{
                \includegraphics[width=0.48\linewidth]{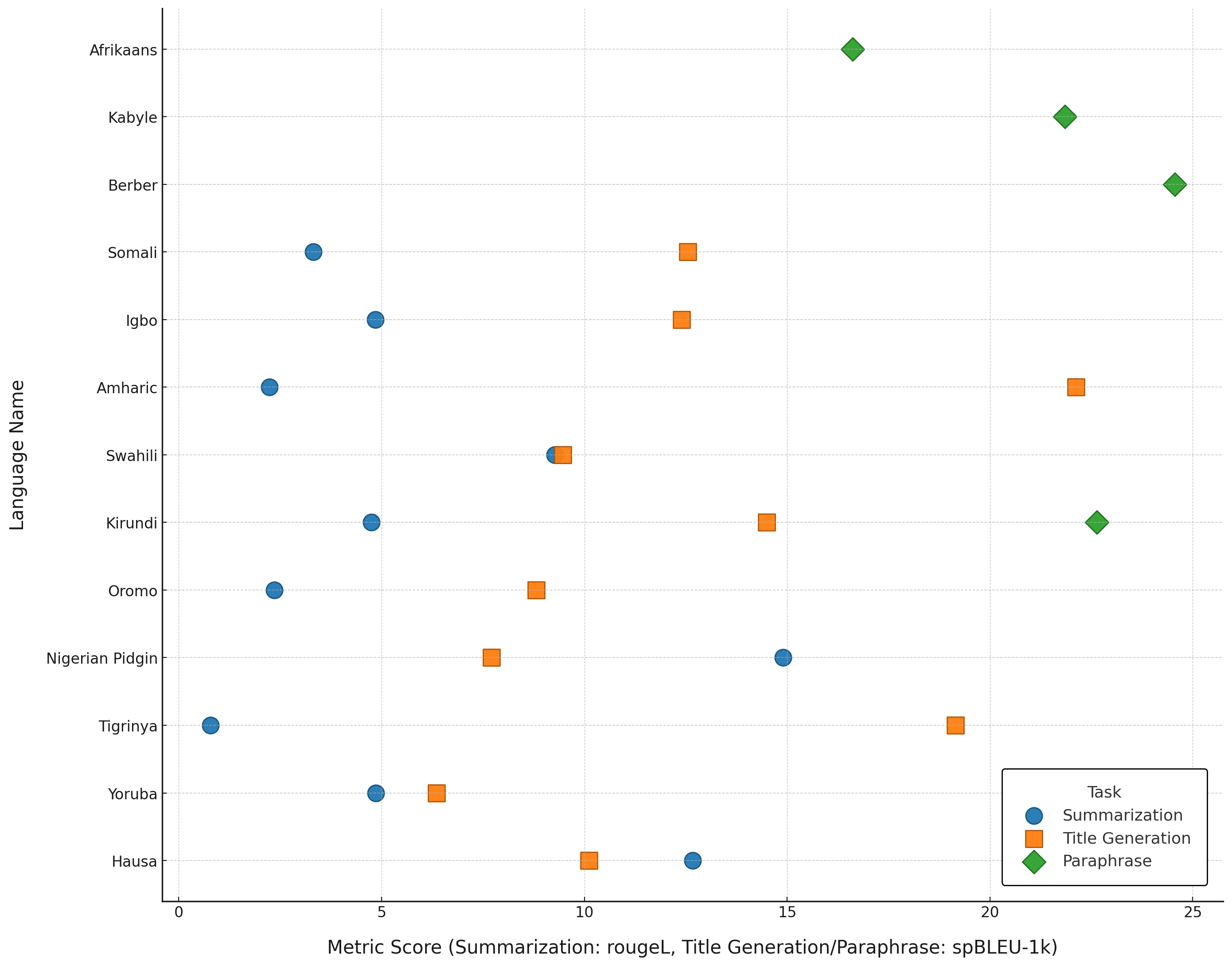}
                \label{fig:preform_sub_1}
            }
            \hfill
            \subfigure[Classification tasks.]{
                \includegraphics[width=0.47\linewidth]{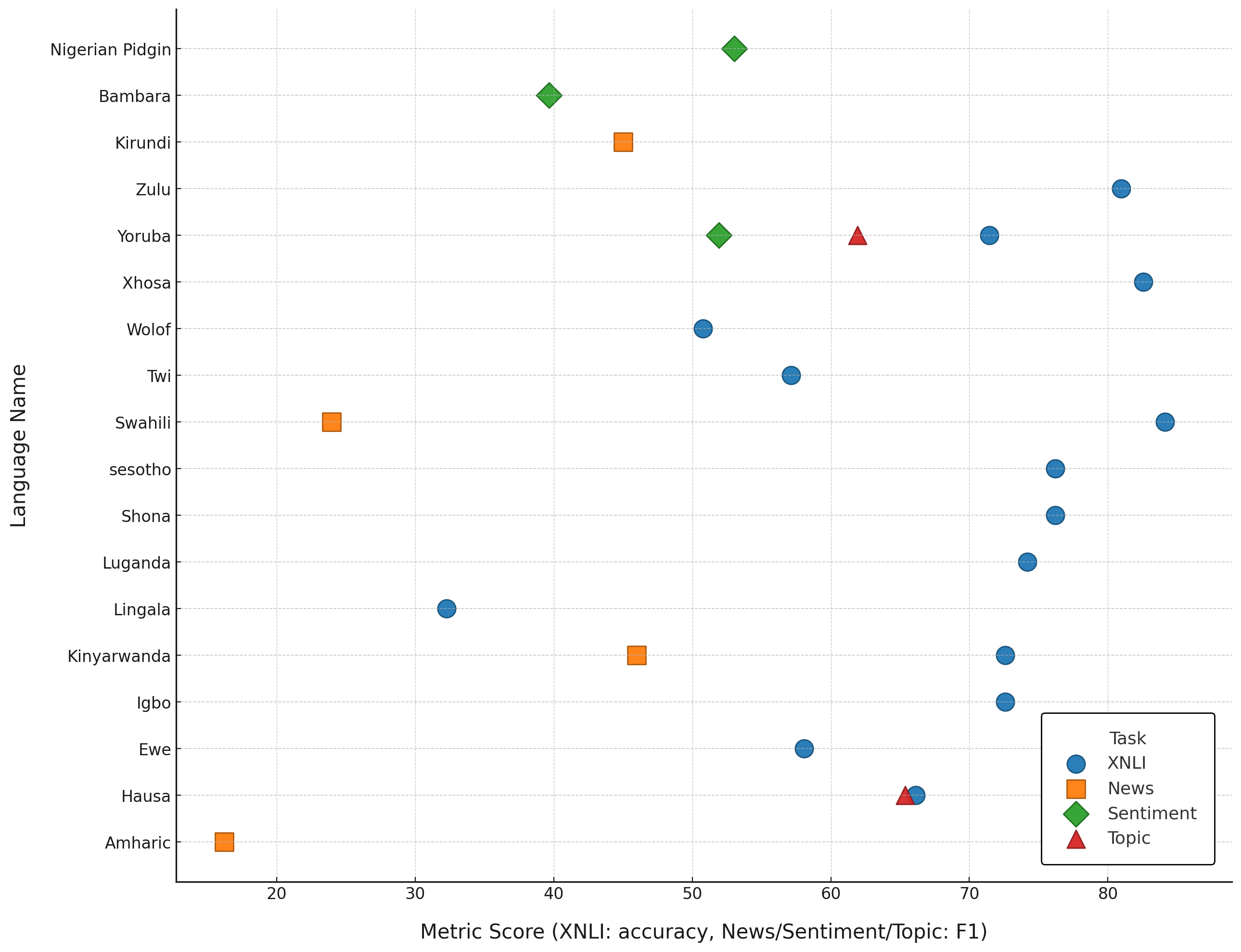}
                \label{fig:preform_sub_2}
            }
            
            \vspace{1em}
            
            \subfigure[MCCR tasks.]{
                \includegraphics[width=0.47\linewidth]{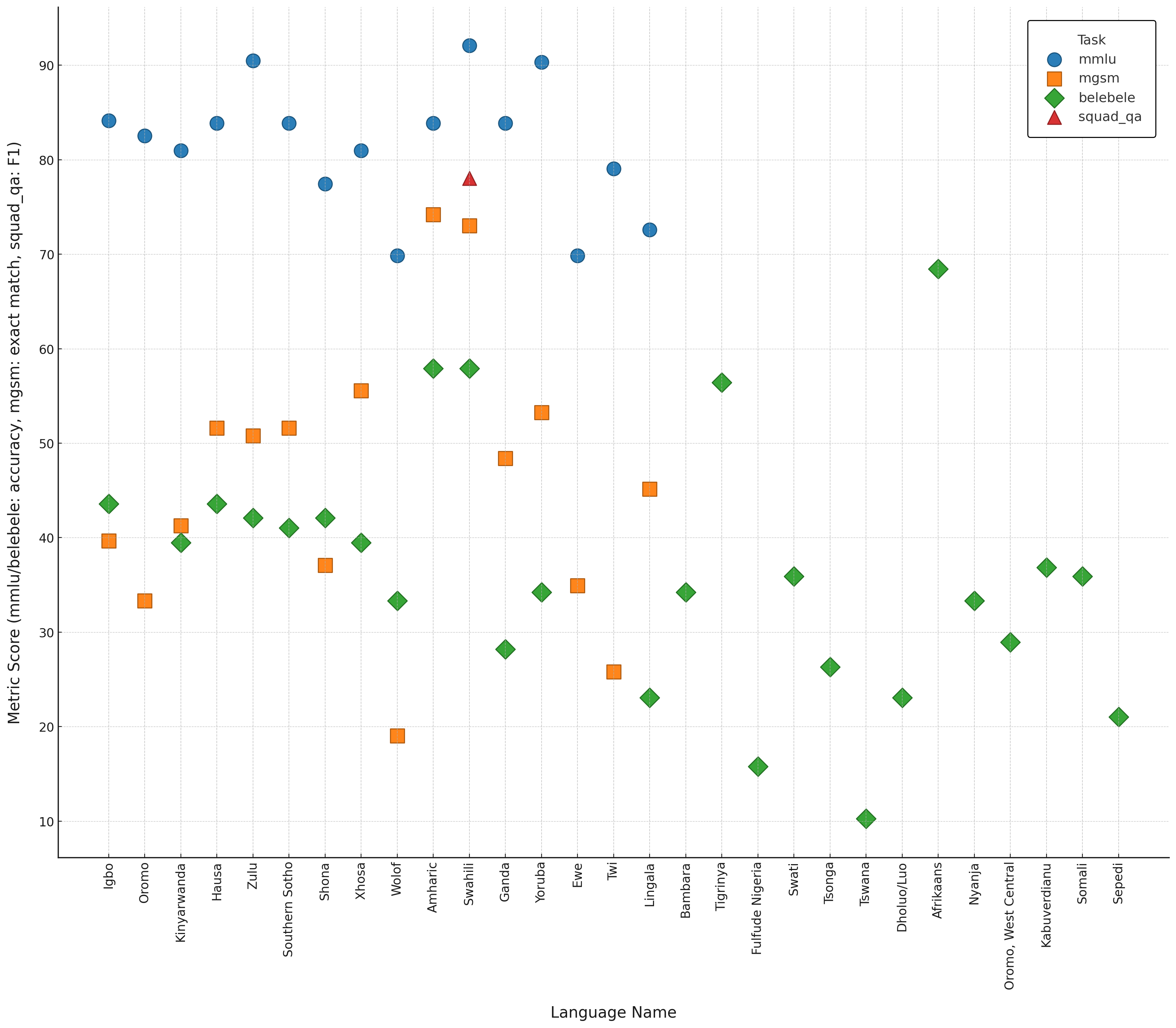}
                \label{fig:preform_sub_3}
            }
            \hfill
            \subfigure[MT tasks grouped.]{
                \includegraphics[width=0.48\linewidth]{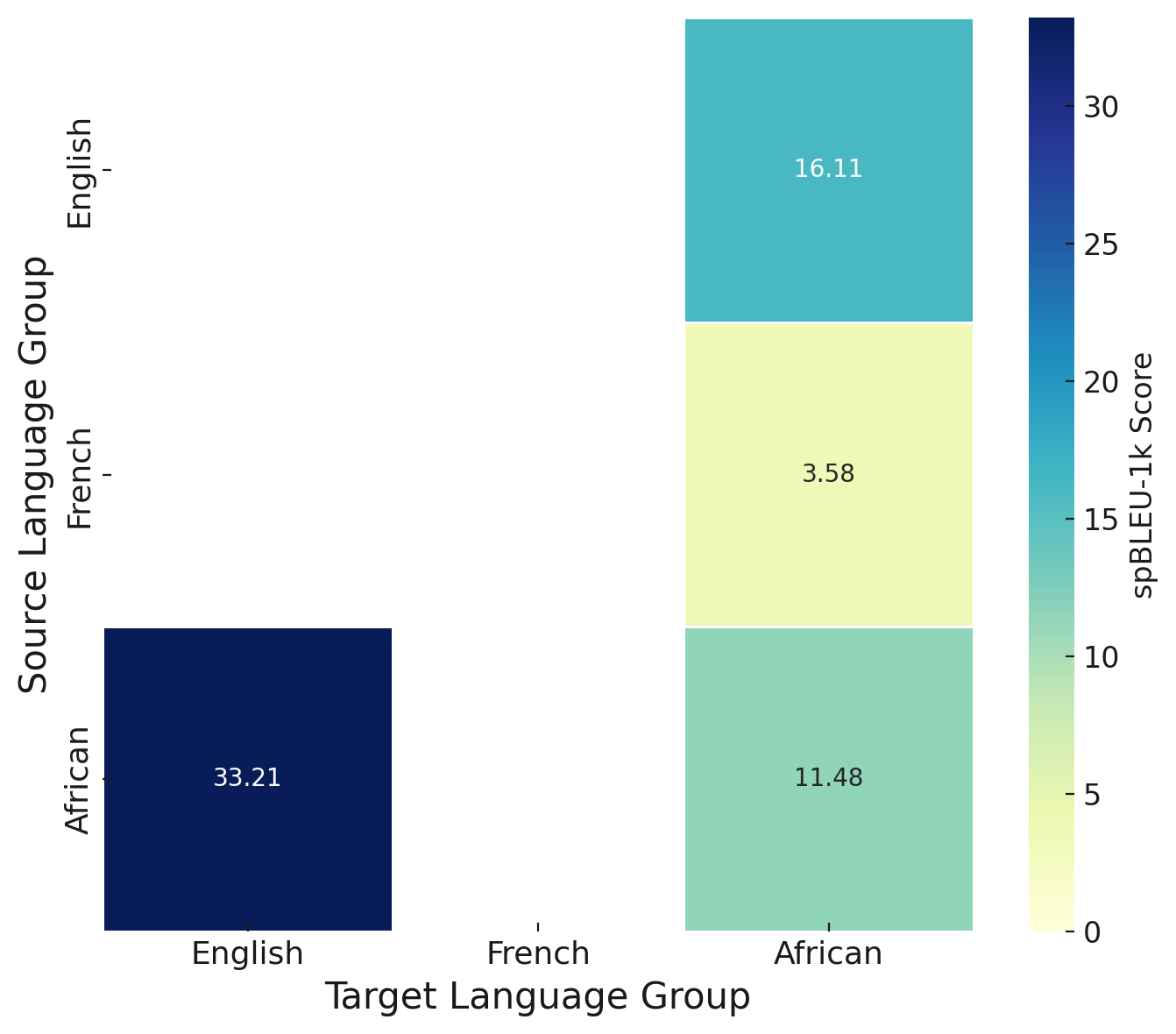}
                \label{fig:preform_sub_4}
            }
        \end{minipage}
    }
    \caption{Distribution of best model performance (Claude-4-Sonnet) on different task clusters across African languages in downstream data.}
    \label{fig:performance_analysis_scatterplot_claude}
\end{figure*}

\paragraph{Model Performance}
Here, we analyze performance of our best model, Claude-4-Sonnet. As shown in Figure~\ref{fig:performance_analysis_scatterplot_claude}, the model faces varying degrees of difficulty when handling African languages. Figure~\ref{fig:preform_sub_1} demonstrates that for generation tasks, even the strongest models struggle to produce high-quality text in African languages, with most scores falling below 15 BLEU/ROUGE. In contrast, Figure~\ref{fig:preform_sub_2} shows that while many languages achieve relatively strong classification performance (over 80\% accuracy), a substantial portion still falls into the 60\%–80\% range, underscoring a clear performance gap. Finally, Figures~\ref{fig:preform_sub_3} and~\ref{fig:preform_sub_4} reveal that translation into English is notably easier than translation out of English or French into African languages, where BLEU scores drop from an average of around $19$ down to nine or $11$, respectively. This finding suggests that even the best model struggles more with producing fluent, faithful output in African languages than with understanding them—underscoring the urgent need for better data and model development tailored to these languages. 

Apart from our empirical analyses, the limited availability and diversity of datasets for African languages in NLP benchmarks has significant implications for developing robust language technologies. \textit{Limited Model Performance and Generalization} – Models trained on a single dataset per language risk overfitting to specific domains or styles, reducing their ability to generalize across different contexts, which is especially problematic for African languages with high dialectal variation. \textit{Bias Toward Well-Resourced Languages} – Languages with more datasets naturally perform better in multilingual models, reinforcing disparities and leading to poor performance for underrepresented African languages. \textit{Restricted Downstream Applications} – Many NLP applications, such as machine translation and question answering, require diverse datasets; if a language is represented only by language identification data, it cannot support the development of more complex language technologies. \textit{Challenges in Transfer Learning} – Transfer learning techniques, like multilingual fine-tuning, rely on cross-lingual similarities and sufficient data; when African languages suffer from sparse training data, they are unable to effectively leverage knowledge transfer from resource-rich languages. We also analyze the performance of Command-A in Figure~\ref{appdx_fig:performance_analysis_scatterplot}.


\section{Recommendations}\label{sec:rec} 
Policy determines how languages are used across various domains, influencing their visibility and viability in digital spaces. Currently, the limited online presence of many Indigenous African languages restricts the availability of training data, which is crucial for building effective NLP models. We provide the following recommendations:

\noindent\textbf{Policy matters.} Governments and policymakers should enact national language policies that mandate the inclusion of Indigenous and underrepresented languages across digital infrastructure, education, and media. Countries with multilingual populations should adopt policies that support the systematic documentation, transcription, and digitization of African languages to enhance their presence in AI applications. Such efforts will enrich the diversity of available data, paving the way for the development of more robust and inclusive models.

\noindent\textbf{Expand data collection and annotation.} Given that only $45$ of the $517$ languages in the benchmark have more than one dataset based on our collection, it is essential to increase the availability of high-quality data for underrepresented African languages across a broad range of downstream tasks. Moreover, prioritizing community-driven data annotation efforts will be crucial in ensuring both linguistic diversity and accuracy, especially for languages that lack institutional support.

\noindent\textbf{Develop contextually relevant and culturally accurate datasets.} Many existing African language datasets are derived from translations of high-resource languages, which can introduce misalignments and fail to capture authentic linguistic expressions. To improve NLP performance, datasets should be created using native speakers and sourced from real-world interactions, literature, and media in African languages. This will help preserve linguistic authenticity and avoid the overreliance on loanwords or structures that do not naturally occur in the language. Specialized corpora should also be developed to capture dialectal variations and culturally specific language use.

\section{Conclusion}\label{sec:conc}
In this work, we set out to unravel how current African language policies, by dictating data availability and accessibility, directly influence model performance across the continent’s diverse linguistic spectrum. To achieve this, we conducted a comprehensive empirical evaluation of leading large language models (LLMs) on \ourbenchmark, an extensive benchmark assembled primarily from existing datasets. \ourbenchmark~consists of $30$ publicly accessible datasets, covering $16$ classification, generation, MCCR, and token level tasks. We also introduced the first publicly available leaderboard for African languages. Our evaluations revealed critical gaps in task performance and language coverage, underscoring pronounced disparities rooted in policy-induced data inequities. Moreover, our findings highlight the importance of integrating theoretical insights on language policies with empirical evaluations to inform forward-thinking policy reforms and promote inclusive data practices. We believe that \ourbenchmark~will serve as essential benchmarking standards, catalyzing future research and development efforts aimed at bridging the digital divide and fostering linguistic diversity in artificial intelligence for African communities.

\section{Limitations}\label{sec:limits}
We identify the following limitations for our work:
\begin{enumerate}
\item \textbf{Data and Data Availability Bias.} Our benchmark relies on publicly available datasets, which inherently reflect existing biases in language documentation and institutional support. Many Indigenous African languages remain severely underrepresented due to historical policies that prioritize dominant languages. As a result, our findings may not fully capture the linguistic diversity of the continent.

\item \textbf{Task Scope.} The evaluation primarily focuses on tasks where data is available, such as language identification and text classification. However, the scarcity of datasets for complex NLP tasks (e.g., machine translation, question answering) limits our ability to assess model performance in broader applications. Future efforts should address this gap by developing more comprehensive datasets across a wider range of tasks.

\item \textbf{Model Performance Constraints.} While we analyze the impact of data scarcity on model effectiveness, our study does not explore architectural modifications or fine-tuning strategies that could improve performance for low-resource languages. Further research is needed to examine whether adaptive techniques, such as multilingual pretraining with targeted data augmentation, could mitigate these challenges.

\item \textbf{Policy Implementation Challenges.} Our recommendations for policy reforms and inclusive data practices require significant institutional commitment and resources. While we outline actionable steps, their adoption depends on the willingness of governments, research institutions, and technology developers to prioritize African language inclusion in NLP.

\end{enumerate}

Despite these limitations, our work underscores the urgency of addressing data inequities and advocating for policies that foster greater linguistic diversity in AI development.

\section{Ethical Considerations}\label{sec:ethics}
We make the following ethics-related statements about our work:
\begin{enumerate}

\item Our research aims to investigate the progress of NLP for African languages, addressing historical language policies that have contributed to data inequities. By improving representation in NLP, we align with broader efforts to foster linguistic diversity in AI, ensuring that African languages are not sidelined in technological advancements.

\item  The datasets used in our benchmark are derived from publicly available sources, yet their existence is shaped by historical and contemporary policies that prioritize certain languages over others. We acknowledge that the digital presence of African languages is not merely a technical issue but a policy-driven reality that influences which languages receive institutional support and investment.

\item Although we do not develop models in this work, our findings highlight the impact of policy-induced data disparities on model performance. Addressing these challenges requires policy interventions that support multilingual data collection, equitable resource allocation, and ethical considerations in dataset creation.

\item Proper attribution to dataset authors is not just an academic necessity but a policy imperative for transparency and recognition. To mitigate concerns about data ownership and ethical usage, we provide a publicly accessible reference file citing all datasets included in our benchmark. We strongly encourage researchers and policymakers to acknowledge these sources, reinforcing the importance of ethical and inclusive data practices.

\end{enumerate}

\section*{Acknowledgments}\label{sec:acknow}
We acknowledge support from Canada Research Chairs (CRC), CLEAR Global funding from the Gates Foundation, the Natural Sciences and Engineering Research Council of Canada (NSERC; RGPIN-2018-04267), the Social Sciences and Humanities Research Council of Canada (SSHRC; 895-2020-1004; 895-2021-1008), Canadian Foundation for Innovation (CFI; 37771), Digital Research Alliance of Canada,\footnote{\href{https://alliancecan.ca}{https://alliancecan.ca}} and UBC ARC-Sockeye.\footnote{\href{https://arc.ubc.ca/ubc-arc-sockeye}{https://arc.ubc.ca/ubc-arc-sockeye}} We also thank Clémentine Fourrier and Nouamane Tazi from HuggingFace for providing feedback on this work. The findings and conclusions
contained within this work those of the authors and do not necessarily reflect positions or policies of any supporters.


\bibliography{anthology,custom}
\bibliographystyle{acl_natbib}

\appendix
\clearpage
\appendixpage
\addappheadtotoc
\setcounter{table}{0}
\setcounter{figure}{0}
\renewcommand{\thetable}{\Alph{section}\arabic{table}}
\renewcommand{\thefigure}{\Alph{section}\arabic{figure}}

\section{Sahara Task Clusters}\label{appendix: task_cluster_detail}
\subsection{Text Classification}
\paragraph{Cross-Lingual Natural Language Inference.}  This task involves the classification of 2 given sentences—a premise and a
hypothesis into entailment, neutral, or contradiction semantic relation \cite{adelani2024irokobench}. \citet{adelani2024irokobench} introduced AfriXNLI, a benchmark for evaluating 15 African Languages on Natural Language Inference (NLI). The languages were translated from the English portion of XNLI \cite{conneau2018xnli}.

\paragraph{Language Identification.} We include language identification datasets across $518$ African languages \cite{adebara2022afrolid}. The dataset includes data from multiple domains including education, government, health care, news, and religion. Additional statistics of the data is available in Table~\ref{tab:data_stats}. We include language identification tasks to investigate \textbf{(1.)} the capabilities of language models in correctly generating the appropriate language names of an input text and \textbf{(2.)} Correctly generate an additional sentence in a specific language. We include these two approaches because many languages have multiple names \cite{chen-etal-2024-fumbling} and a LM's inability to appropriately name a language does not implicate a lack of information about the language. 

\paragraph{News Classification.} This task involves automatically categorizing news articles into predefined categories or topics based on their content. The goal is to leverage machine learning and NLP techniques to understand and organize large volumes of news data, making it easier for users to access relevant information. News classification plays a crucial role in information retrieval, content recommendation, and other applications that involve organizing and categorizing textual data. The news classification cluster consists of data from four languages including Amharic~\cite{azime2021amharic}, Kinyarwanda~\cite{niyongabo-etal-2020-kinnews}, Kirundi~\cite{niyongabo-etal-2020-kinnews}, and Swahili~\cite{davis_david_2020_4300294, davis_david_2020_5514203}. Table~\ref{tab:data_stats} provides additional details of the data in this cluster. 

\paragraph{Sentiment Analysis.} Sentiment analysis is crucial in gaining insights into public opinion, customer feedback, and user sentiments across various platforms. The sentiment analysis cluster consists of the Bambara Sentiment dataset \cite{diallo2021bambara}, YOSM--a Sentiment Corpus for Movie Reviews in Y{o}r\`{u}b\'{a}~\cite{shode_africanlp}, and the Nigerian Pidgin sentiment dataset~\cite{oyewusi2020semantic}, respectively. Further details about the statistical composition of this data is available in Table~\ref{tab:data_stats}.

\paragraph{Topic Classification.} The primary goal for this task is to automatically categorize the content of the text based on its theme. Topic classification is widely used in various applications, such as document organization, content recommendation, and information retrieval. We include topic classification datasets for Y{o}r\`{u}b\'{a} and Hausa~\cite{hedderich-etal-2020-transfer}.

\subsection{Text Generation}

\paragraph{Machine Translation.} A robust benchmark for MT tasks is essential. \ourbenchmark, in its quest to effectively assess machine translation for African languages, selectively incorporates datasets containing diverse African languages into its benchmark. Specifically, \ourbenchmark~includes datasets such as Pidgin-UNMT\footnote{\href{https://github.com/keleog/PidginUNMT}{PidginUNMT GitHub Link}}~\cite{ogueji2019pidginunmt}, Afro-MT\footnote{\href{https://github.com/machelreid/afromt}{AfroMT GitHub Link}}~\cite{reid-etal-2021-afromt}, Lafand-MT~\cite{adelani-etal-2022-thousand}, SALT\footnote{\href{https://github.com/SunbirdAI/salt}{SALT GitHub Link}} \cite{akera2022salt}, and HornMT, thus encompassing a total of $45$ language pairs.

\paragraph{Paraphrase.} Paraphrasing is a core task in NLG that revolves around the comprehension and generation of text that conveys the same meaning while possibly exhibiting differences in structure, phrasing, or style~\cite{10.1162/tacl_a_00542}. \ourbenchmark~incorporates the TaPaCo dataset~\cite{scherrer_yves_2020_3707949}, which is publicly accessible and covers a total of $73$ languages, including four of African origin. These African languages consist of Kirundi, Amazigh, a macro-language Berber, and Afrikaans.

\paragraph{Summarization.} Summarization is the task of creating a condensed version of a text that retains its core ideas and essential information~\cite{nallapati-etal-2016-abstractive}. In the context of \ourbenchmark, our focus primarily centers on a subset of the abstractive summarization dataset, XL-SUM~\cite{hasan-etal-2021-xl}, which includes several African languages: Amharic, Hausa, Igbo, Kirundi, Oromo, Pidgin, Somali, Swahili, Tigrinya, and Yoruba. Additionally, we have extended our dataset collection efforts by scraping non-overlapping summarization datasets from BBC and Voice of Africa (VoE) websites, specifically targeting three African languages: Hausa, Ndebele, and Swahili.

\paragraph{Title Generation.} The task of title generation in NLG involves generating a concise and meaningful title or headline for a given article. This task shares its dataset source with summarization, drawing from XL-SUM, and also includes dataset extensions from BBC and VoE with focus on enabling zero-shot title generation.\\

\subsection{Multiple-Choice, Comprehensive and Reasoning}

This task cluster majorly addresses \emph{Question Answering} type questions. The QA tasks in this cluster are more mathematical-reasoning and reading-comprehension-oriented.

\paragraph{Mathematical Word Problems.} \cite{adelani2024irokobench} introduced AfriMGSM, a QA dataset that contains human-written grade school mathematical word problems. Their dataset covers 15 African Languages, and the translation of each question to English for each language for in-language evaluation of LLMs. In this task, given a word problem, the LLM is expected to return a numerical value as output.

\paragraph{General Knowledge QA.} \cite{adelani2024irokobench} introduced AfriMMLU, a multi-choice QA dataset covering 5 areas of knowledge in 15 African Languages. The knowledge areas are high-school
geography, high-school microeconomics, elementary mathematics, international law), and global facts. In this task given a question in any of the aforementioned areas, and 4 options, the LLM is expected to choose the right option.

\paragraph{Reading Comprehension QA.} In this task, the LLMs are evaluated based on their ability to understand information in a given article. \cite{belebele_2024} introduced a parallel reading comprehension dataset covering 122 languages, 23 of which are native to Africa. The dataset comprises of short paragraphs accompanied by multi-choice Questions, whose answers can be inferred from accompanying paragraphs.

\paragraph{Context-based Question-Answering.} QA represents another vital facet of NLG, focusing on a model's capability to provide answers based on integrated knowledge. \ourbenchmark~relies on the gold passage dataset from the multilingual TYDIA\footnote{\href{https://github.com/google-research-datasets/tydiqa}{TyDiQA GitHub Link}} QA resource~\cite{clark-etal-2020-tydi}. This dataset offers concise questions with precisely one corresponding answer, enabling rigorous evaluation of QA performance.

\subsection{Tokens-level Classification}

\paragraph{Named Entity Recognition.} We incorporate NER datasets across $30$ languages. We use Distance Supervision NER (DS NER) Data \cite{hedderich-etal-2020-transfer}, MasakhaNER~\cite{adelani-etal-2022-masakhaner}, WikiAnn~\cite{rahimi-etal-2019-massively}, Yoruba-Twi NER~\cite{alabi2020massive}, and multiple NER data from \href{https://repo.sadilar.org/handle/20.500.12185/7}{SADiLaR}. Additional details about the datasets are available in Table~\ref{tab:data_stats}.

\paragraph{Phrase Chunking.} This task involves identifying and grouping together consecutive words or tokens into meaningful syntactic units, known as phrases. These phrases can include noun phrases, verb phrases, prepositional phrases, and other grammatical structures. The goal of phrase chunking is to analyze the grammatical structure of a sentence and extract higher-level syntactic units for better understanding. The phrase chunking cluster consists of data for ten Indigenous languages of South Africa (see Table \ref{tab:data_stats}). The data has annotations for noun, verb, adjective, adverbial, and prepositional phrase chunks. Words not belonging to these phrase types are labelled with the tag \textit{O}. 

\paragraph{Part of Speech Tagging.} This classification task involves assigning grammatical categories or labels (such as noun, verb, adjective) to each word in a sentence. The primary goal of POS tagging is to analyze the syntactic structure of a sentence and understand the role of each word within it. POS tagging is crucial for various NLP applications, as it provides a foundation for more advanced linguistic analysis. We include datasets for Igbo taken from IgboNLP~\cite{10.1145/3314942}.

\section{Benchmark}
Table~\ref{tab:benchmarkcompare}~shows the comparison between \ourbenchmark~with other benchmarks.
\begin{table*}[h]
\scriptsize
\centering
\resizebox{0.85\textwidth}{!}{%
\begin{tabular}{clp{3.5cm}ccc}
\toprule
\textbf{Category} & \textbf{Benchmark} & \textbf{Reference}& \textbf{Lang/Total} & \textbf{Tasks} \\ \toprule
\multirow{9}{*}{\rotatebox[origin=c]{90}{\textbf{Multilingual}}} & FLoRES200 & \cite{nllb2022} & 52/200 & 1 \\
& GEM\textsubscript{v1} & \cite{gehrmann-etal-2021-gem} & 10/52 & 5 \\
& GEM\textsubscript{v2} & \cite{gehrmann-etal-2021-gem} & 10/52 & 9\\
& NLLB M.D. & \cite{nllb2022} & 2/8 &  1 \\
& NLLB S.D. & \cite{nllb2022} & 2/8 & 1 \\
& SIB-200 & \cite{adelani-etal-2024-sib} & 46/200 & 1\\
& Toxicity200 & \cite{nllb2022}& 50/200 & 1 \\
& XGLUE & \cite{liang-etal-2020-xglue} & 1/19 & 3\\
& XTremE & \cite{hu2020xtreme} & 2/40 & 4\\
\midrule
\multirow{9}{*}{\rotatebox[origin=c]{90}{\textbf{Language-specific}}} & BanglaNLP\textsubscript{v1} & \cite{bhattacharjee-etal-2023-banglanlg} & 1 & 6 \\
& CLUE & \cite{xu2020clue} & 1 & 6 \\
& CUGE & \cite{yao2021cuge} & 1 & 8 \\
& Dolphin & \cite{nagoudi-etal-2023-dolphin} & 1 & 13 \\
& FLUE & \cite{le2020flaubert} & 1 & 5 \\
& IndoNLU & \cite{wilie2020indonlu} & 1 & 5 \\
& IndoNLG & \cite{guntara-etal-2020-benchmarking} & 1 & 6 \\
& JGLUE & \cite{kurihara2022jglue} & 1 & 3 \\
& KorNLU & \cite{ham2020kornli} & 1 & 2 \\
& LOT & \cite{10.1162/tacl_a_00469} & 1 & 4 \\
& ORCA & \cite{elmadany-etal-2023-orca} & 1 & 7 \\
& PhoMT & \cite{doan-etal-2021-phomt} & 1 & 1 \\
\midrule
\multirow{9}{*}{\rotatebox[origin=c]{90}{\textbf{African}}} & AfriSenti & \cite{muhammad-etal-2023-afrisenti}& 14 & 1 \\
& AfroMT & \cite{reid21afromt} & 8/8 & 1 \\
& AfroNLG & \cite{adebara2024cheetah} & 517/517 & 6 \\ 
& AfroNLU & \cite{adebara-etal-2023-serengeti} & 28/28 & 7 \\
& \href{https://github.com/asmelashteka/HornMT}{Horn-MT} & $-$ & 3/3 & 1 \\
& Iroko-Bench & \cite{adelani2024irokobench} & 16/16 & 3\\
& Mafand-MT & \cite{adelani-etal-2022-thousand} & 17/17 & 1 \\ 
& Menyo-20k & \cite{adelani-etal-2021-effect} & 1/1 & 1 \\
& Naija-senti & \cite{muhammad-etal-2022-naijasenti} & 4/4 & 1 \\ \cmidrule{2-5}
&\includegraphics[scale=0.014]{Images/sahara_logo.png}  \ourbenchmark & ours & 517/517 & 16 \\ \bottomrule
\end{tabular}%
}
\caption{A Comparison of \ourbenchmark~with other Benchmarks. \textbf{M.D:} Multilingual domain, \textbf{S.D:} Seed Data.}
\label{tab:benchmarkcompare}
\end{table*}

\section{Few-shot prompt example}
\begin{figure*}[h]
  \centering
  \includegraphics[width=\textwidth]{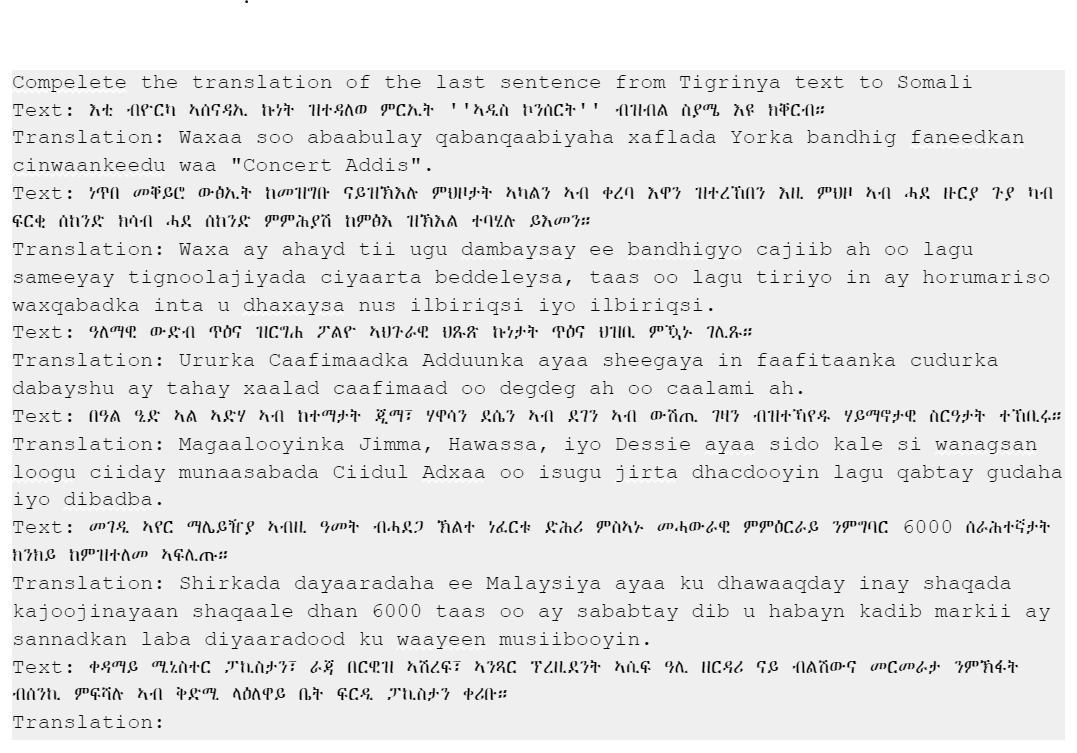}
  \caption{Few-shot prompting example for the machine translation task. The prompt provides five examples of the specific task for context and alignment. }
  \label{fig:few-shot}
\end{figure*}

Figure~\ref{fig:few-shot} shows an example for MT task. 

\section{Dataset Description}
Table~\ref{tab:description} describes all datasets included in \ourbenchmark.

\begin{table*}[!ht]
    \centering  
    \small
    \resizebox{0.85\textwidth}{!}{
        \begin{tabularx}{\textwidth}{XXX}
        \toprule
        Dataset & Brief Description & African Language(s) \\
        \midrule
        
        MasakhaNER \cite{Adelani2021MasakhaNERNE}, \cite{Adelani2022MasakhaNER2A}  & Curated for Named Entity Recognition for several African Languages  & amh, bam, bbj, ewe, fon, hau, ibo, kin, lug, luo, mos, nya, pcm, sna, swa, tsn, twi, wol, xho, yor, zul \label{masakhanener}\\ \midrule
        MasakhaPOS \cite{Dione2023MasakhaPOSPT} & Curated from News sources and publicly available datasets for Part-Of-Speech Tagging & bam, bbj, ewe, fon, hau, ibo, kin, lug, luo, mos, nya, pcm, sna, swa, tsn, twi, wol, xho, yor, zul   \\ \midrule
        MAFAND-MT \cite{adelani-etal-2022-thousand} & African news corpus for Machine Translation  &  bam, bbj, ewe, fon, hau, ibo, lug, luo, mos, swa, tsn, twi, wol, xho, yor, zul  \\ \midrule
        XL-Sum \citet{hasan-etal-2021-xl} & Curated for Summarization and Title Generation &  amh, ibo, orm, run, swa, yor, hau, pcm, som, tir \\ \midrule
        YorubaSenti \cite{yorubasenti} & Curated from comments on movies on YouTube
 & yor \\ \midrule
         TaPaCo \cite{scherrer_yves_2020_3707949}  & Developed to provide example sentences and translations for particular linguistic constructions and words  & afr, ber, run , zgh \\ \midrule
         TyDi QA \cite{clark-etal-2020-tydi} & Curated for Information-Seeking QA tasks & swa  \\ \midrule
         HornMT & Parallel MT dataset curated for languages in Horn of Africa  & aaf, amh, orm, som, tir \\ \midrule
         yoruba-twi-ner \cite{alabi-etal-2020-massive} & Curated for evaluating embeddings for low-resource languages & yor, twi\\ \midrule
         KINNEWS and KIRNEWS \cite{niyongabo-etal-2020-kinnews} & Curated for multi-class classification of news articles & kin, run \\ \midrule
         AmharicNews \cite{azime2021amharic}& Curated for text classification from news articles &amh \\ \midrule
         SwahiliNews v0.2 \cite{davis_david_2020_4300294}& Curated for text classification from news articles & swa\\ \midrule
         Bambara \cite{diallo2021bambara} & Curated for Sentiment Analysis through web crawling & bam \\ \midrule
         IgboNLP\cite{igbonlp} & Contains text corpus, POS tagset and POS-tagged subcorpus & ibo \\ \midrule
         yosm \cite{shode2022yosm} & Curated from movie reviews for Sentiment Analysis & yor \\ \midrule
         pidgin-tweet \cite{oyewusi2020semantic}& Curated for SA from tweets & pcm \\ \midrule
         WikiAnn \cite{pan-etal-2017-cross}, \cite{rahimi-etal-2019-massively}  & & afr, amh, ibo, mlg, kin, som, swa, yor \\ \midrule
         HausaTopic,YorubaTobic \cite{hedderich-etal-2020-transfer} & Curated for Topic Classification & hau, yor \\ \midrule
          SaDiLAR \cite{eiselen-2016-government} & Curated for NER of 10 South African Languages & afr, nbl, nso, sot, ssw, tsn, tso, ven, xho, zul \\ \midrule
         AfriSenti \cite{muhammad-etal-2023-afrisenti} & Curated for African tweet sentiment analysis  & amh, ALG-ara, hau, kin, ibo, MOR-ara, MOZ-por, pcm,form, swa, tir, twi,tso,yor \\ \midrule
         Afro-MT \cite{reid-etal-2021-afromt} & Benchmark for MT of 8 African Languages & afr, bem, lin, run, sot, swa, xho, zul \\ \midrule
         PidginUNMT \cite{ogueji-etal-2021-small} & & pcm\\ \midrule
         SALT \cite{akera2022salt}& Curated for Parallel MT of 5 Ugandan Languages & ach, lgg, lug, nyn, teo   \\ \midrule
         Cheetah \cite{adebara2024cheetah} & Summarization test set & hau, nde, swa \\ \midrule
         Cheetah \cite{adebara2024cheetah} & Title generation test set &  amh, gag (zero-shot), hau, ibo, pcm, som, swa, tir, yor, kin (zero-shot), afr, mlg (zero-shot), orm, nde (zero-shot), sna(zero-shot)\\ \midrule
         NCHLT-NER \cite{eiselen-2016-government} & Named Entity Recognition & afr, nbl, nso, sot, ssw, tsn, tso, ven, xho, zul  \\ \midrule
         BeleBele \cite{belebele_2024} & Reading Comprehension MCQA & afr
amh, bam, fuv, gaz, hau, ibo,kea,kin, lin, lug, luo,nso, nya, sna ,som, sot, ssw, swh, tir, tsn, tso, wol, xho, yor,zul  \\ \midrule
IrokoBench \cite{adelani2024irokobench} & MGSM,MMLU,XNLI & amh, ewe, hau, ibo,kin, lin, lug, orm, sna ,sot,  swa, tir, wol, xho, yor,zul  \\ \bottomrule
        \end{tabularx}%
    }
    \caption{A description of all datasets included in \textit{Sahara}. \textbf{ALG:} Algerian, \textbf{MOR:} Moroccan, \textbf{MOZ:} Mozambican.}
    \label{tab:description}
\end{table*}

\begin{table*}[]
\centering
\resizebox{0.55\textwidth}{!}{%
\begin{tabular}{@{}l|l|l|r@{}}
\toprule
\textbf{Task} & \textbf{Source} & \textbf{Language} & \textbf{Train/ Dev/ Test} \\ \midrule
\multirow{26}{*}{MT} & \multirow{15}{*}{\cite{adelani-etal-2022-thousand}} & eng-hau & 5.9K/ 1.3K/ 1.5K \\
&  & eng-ibo & 6.9K/ 1.5K/ 1.4K \\
 &  & eng-lug & 4.1K/ 1.5K/ 1.5K \\
 &  & eng-pcm & 4.8K/ 1.5K/ 1.6K \\
 &  & eng-swa & 30.8K/ 1.8K/ 1.8K \\
 &  & eng-tsn & 2.1K/ 1.3K/ 1.5K \\
 &  & eng-twi & 3.3K/ 1.3K/ 1.5K \\
 &  & eng-yor & 6.6K/ 1.5K/ 1.6K \\
 &  & eng-zul & 3.5K/ 1.5K/ 1.0K \\
 &  & fra-bam & 3.0K/ 1.5K/ 1.5K \\
 &  & fra-bbj & 2.2K/ 1.1K/ 1.4K \\
 &  & fra-ewe & 2.0K/ 1.4K/ 1.6K \\
 &  & fra-fon & 2.6K/ 1.2K/ 1.6K \\
 &  & fra-mos & 2.5K/ 1.5K/ 1.6K \\
 &  & fra-wol & 3.4K/ 1.5K/ 1.5K \\ \cmidrule(l){2-4} 
 & \multirow{8}{*}{\cite{reid-etal-2021-afromt}} & eng-afr & 25.8K/ 3.2K/ 3.2K \\
 &  & eng-bem & 12.0K/ 1.5K/ 1.5K \\
 &  & eng-lin & 17.7K/ 2.2K/ 2.2K \\
 &  & eng-run & 12.5K/ 1.6K/ 1.6K \\
 &  & eng-sot & 28.8K/ 3.6K/ 3.6K \\
 &  & eng-swa & 28.1K/ 3.5K/ 3.5K \\
 &  & eng-xho & 26.1K/ 3.3K/ 3.3K \\
 &  & eng-zul & 29.1K/ 3.6K/ 3.6K \\ \cmidrule(l){2-4} 
 & \cite{ogueji2019pidginunmt} & eng-pcm & 1.7K/ 0.2K/ 0.2K \\ \cmidrule(l){2-4} 
 & \cite{akera2022salt} & All-pairs$^1$ & 20.0K/ 2.5K/ 2.5K \\ \cmidrule(l){2-4} 
 & \href{https://github.com/asmelashteka/HornMT}{HornMT} & All-pairs$^2$ & 1.7K/ 0.2K/ 0.2K \\ \midrule
\multirow{5}{*} {NER} & \cite{Adelani2021MasakhaNERNE}& See Table \ref{tab:description}  & 41.2K/ 5.1K/ 5.1K \\ \cmidrule(l){2-4} 
& \cite{Adelani2022MasakhaNER2A} & See Table \ref{tab:description} & 41.2K/ 5.1K/ 5.1K \\ \cmidrule(l){2-4}
& \cite{eiselen-2016-government} &  See Table \ref{tab:description} & 1.7M/ 219.7K/ 215.6K \\ \cmidrule(l){2-4}
& \cite{alabi-etal-2020-massive} & See Table \ref{tab:description}  & 20.2K/ 2.8K/ 5.5K \\ \cmidrule(l){2-4}
& \cite{pan-etal-2017-cross} & See Table \ref{tab:description}  & 9.2K/ 9.2K/ 9.4K \\ \midrule 
\multirow{4}{*} {News} & \cite{azime2021amharic}& amh  & 41.2K/ 5.1K/ 5.1K \\ \cmidrule(l){2-4} 
& \multirow{2}{*}{\cite{niyongabo-etal-2020-kinnews}} & kin  & 15.3K/ 1.7K/ 4.3K \\  
& & run  & 3.3K/ 0.4K/ 0.9K \\  \cmidrule(l){2-4} 
&\cite{davis_david_2020_4300294} & swa  & 4.4K/ 0.6K/ 0.6K \\  \midrule 
\multirow{3}{*} {Paraphrase} & \multirow{1}{*}{\cite{scherrer_yves_2020_3707949} }& amh, ber, kab, run   & 22.4K/ 2.8K/ 2.8K \\
& & ber  & 17.6K/ 2.2K/ 2.2K \\  
& & kab & 4.4K/ 0.6K/ 0.6K \\  \midrule
Phrase Chunking & \cite{eiselen-2016-government} & See Table \ref{tab:description} & 107.5K/ 13.0K/ 13.4K \\ \midrule 
POS Tagging & ~\cite{10.1145/3146387, 10.1145/3314942} & ibo & 756.8K/ 94.7K/ 95.0K \\ \midrule
Question Answering & \cite{clark-etal-2020-tydi} & swa & 49.9K/ 0.5K/ n/a \\  \midrule
\multirow{3}{*}{Sentiment Analysis}& \cite{diallo2021bambara} & bam & 2.4K/ 0.3K/0.3K \\ \cmidrule(l){2-4} 
& \cite{oyewusi2020semantic} & pcm & 11.2K/ 1.4K/ 1.4K \\ \cmidrule(l){2-4} 
& \cite{shode2022yosm} & yor & 0.8K/ 0.2K/ 0.5K \\ \midrule
\multirow{12}{*}{Summarization} & \multirow{11}{*} {\cite{hasan-etal-2021-xl}} &  amh & 5.8K/ 0.7K/ 0.7K \\ 
 & & ibo & 4.2K/ 0.5K/ 0.5K \\  
 & & orm & 6.1K/ 0.8K/ 0.8K \\  
  && Rundi & 5.7K/ 0.7K/ 0.7K \\  
 && swa & 7.9K/ 1.0K/ 1.0K \\  
 && yor & 6.4K/ 0.8K/ 0.8K \\  
 && hau & 6.4K/ 0.8K/ 0.8K \\  
 && pcm & 9.2K/ 1.2K/ 1.2K \\  
 && som & 6.0K/ 0.7K/ 0.7K \\  
 && Tigrinya & 5.5K/ 0.7K/ 0.7K \\ 
 \cmidrule{2-4}
 & \cite{adebara2024cheetah}& \textsuperscript{$\star\dagger$} & -/ -/ 0.4K \\ \midrule
\multirow{12}{*}{Title Generation} & \multirow{11}{*} {\cite{hasan-etal-2021-xl}} &  amh & 5.8K/ 0.7K/ 0.7K \\ 
 && ibo & 4.2K/ 0.5K/ 0.5K \\  
 && orm & 6.1K/ 0.8K/ 0.8K \\  
 && run & 5.7K/ 0.7K/ 0.7K \\  
 && swa & 7.9K/ 1.0K/ 1.0K \\  
 && yor  & 6.4K/ 0.8K/ 0.8K \\  
 && hau  & 6.4K/ 0.8K/ 0.8K \\  
 && pcm & 9.2K/ 1.2K/ 1.2K \\  
 && som & 6.0K/ 0.7K/ 0.7K \\  
 && Tigrinya & 5.5K/ 0.7K/ 0.7K \\ 
 \cmidrule{2-4}
 &\cite{adebara2024cheetah}& \textsuperscript{$\star$}  & -/ -/ 5.9K \\ \midrule
\multirow{2}{*}{Topic Classification} & \multirow{2}{*}{\cite{hedderich-etal-2020-transfer}} & hau & 2.0K/ 0.3K/0.6K \\ 
& & yor & 1.3K/ 0.2K/ 0.4K \\ 

 \bottomrule
\end{tabular}%
}
\caption{Statistics of the data in our benchmark. $All-pairs^1$ each have the same size of data. They include ach-eng, ach-lgg, ach-lug, ach-nyn, ach-teo, ach-teo, eng-lgg, eng-lug, eng-nyn, eng-teo, lgg-teo, lug-lgg, lug-teo, nyn-lgg, nyn-lug, and nyn-teo. $All-pairs^2$ are all possible language combinations of aaf, amh, orm, som, tir, eng. \textsuperscript{$\star\dagger$} is a summarization test set including `hau', `nde' (zero-shot), and `swa'. \textsuperscript{$\star$} is a title generation test set across 15 languages: `amh', `gag' (zero-shot), `hau', `ibo', `pcm', `som', `swa', `tir', `yor', `kin' (zero-shot), `afr', `mlg' (zero-shot), `orm', `nde' (zero-shot), `sna'(zero-shot).}
\label{tab:data_stats}
\end{table*}



\section{Data-Performance-Policy Analysis}\label{appdx_sec:analysis}

\begin{figure*}[!ht]
    \centering
    \resizebox{0.95\textwidth}{!}{ 
        \begin{minipage}{\linewidth} 
            \centering
            \subfigure[Generation tasks.]{
                \includegraphics[width=0.48\linewidth]{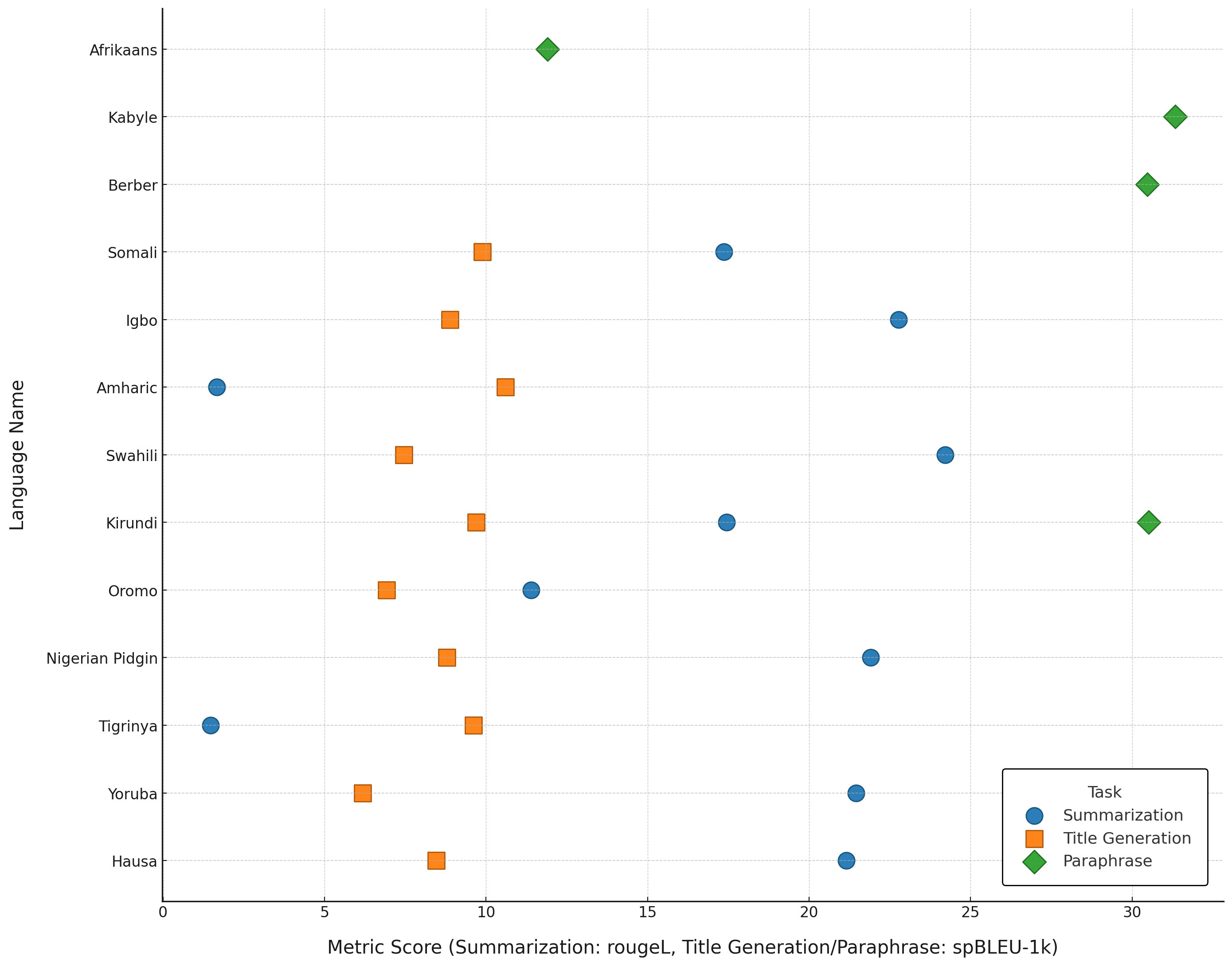}
                \label{fig:preform_sub1}
            }
            \hfill
            \subfigure[Classification tasks.]{
                \includegraphics[width=0.47\linewidth]{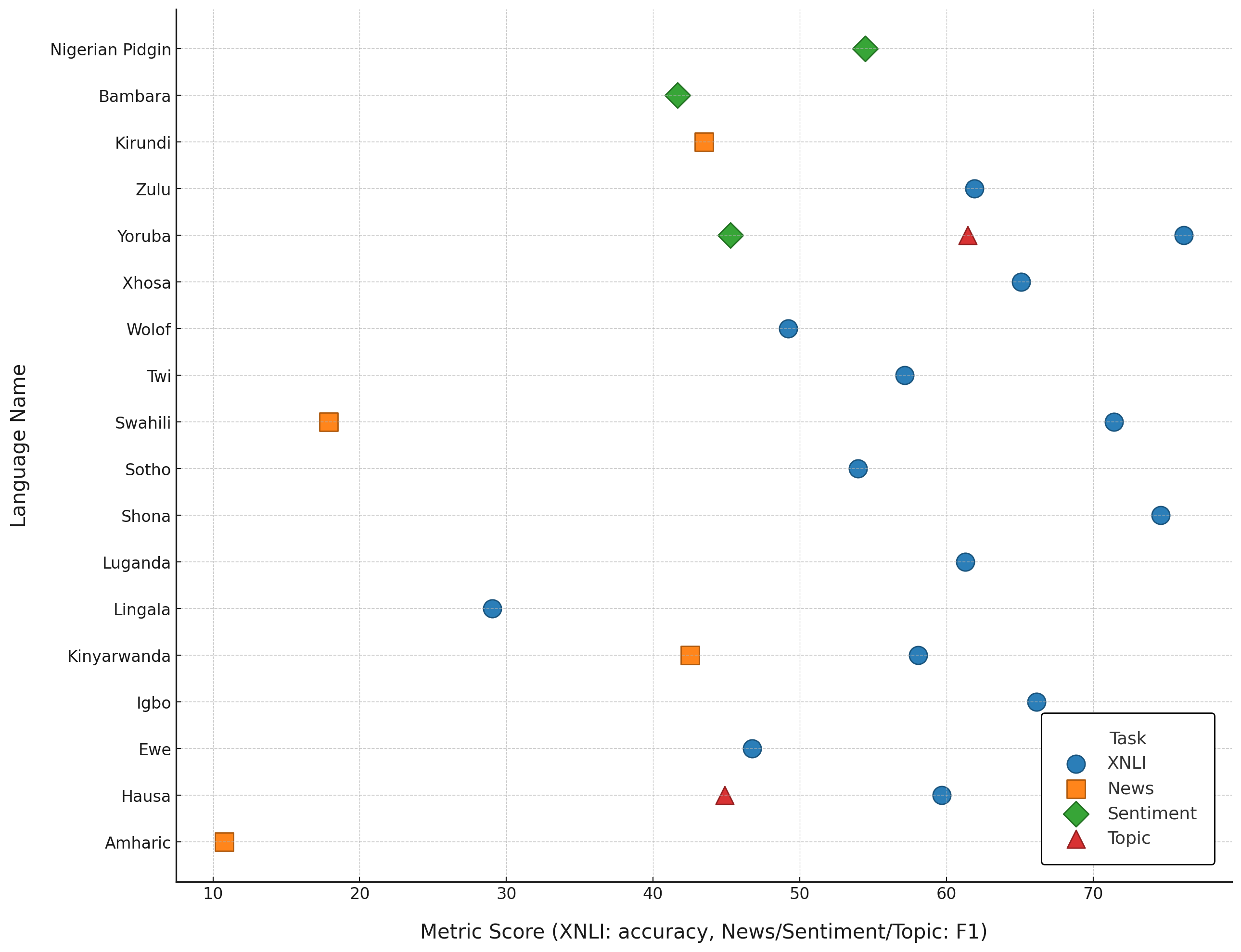}
                \label{fig:preform_sub2}
            }
            
            \vspace{1em}
            
            \subfigure[MCCR tasks.]{
                \includegraphics[width=0.47\linewidth]{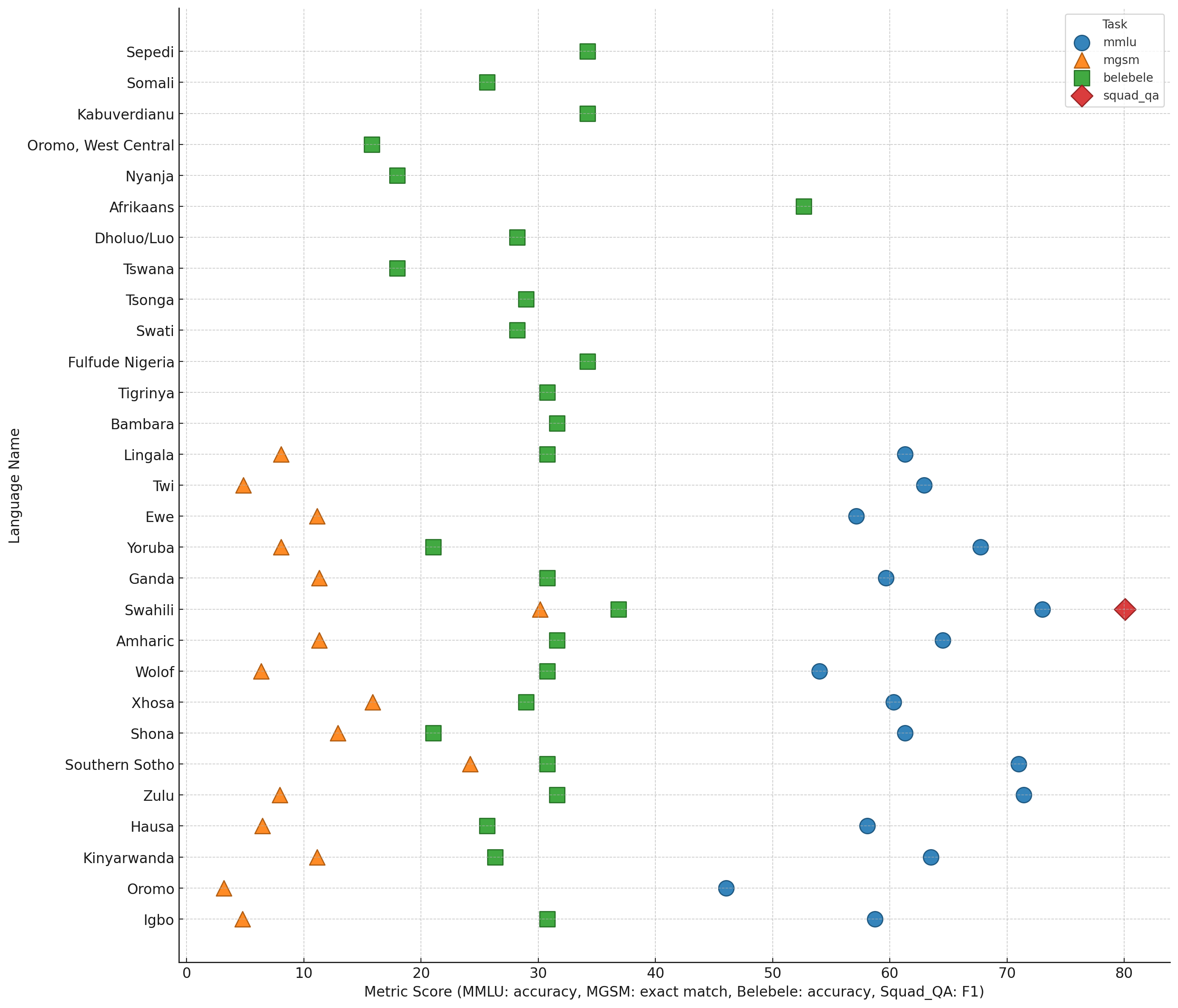}
                \label{fig:preform_sub3}
            }
            \hfill
            \subfigure[MT tasks grouped.]{
                \includegraphics[width=0.48\linewidth]{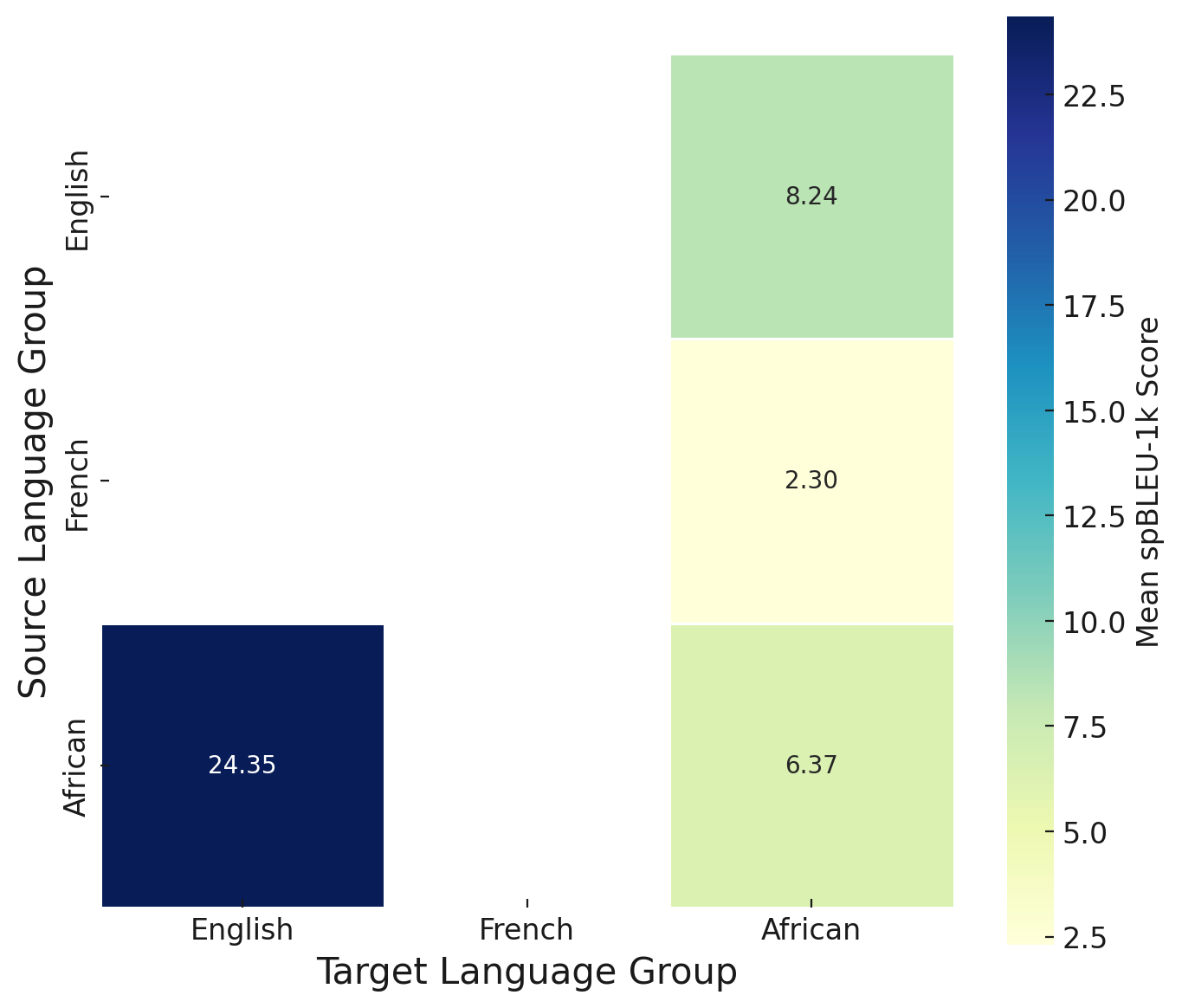}
                \label{fig:preform_sub4}
            }
        \end{minipage}
    }
    \caption{Distribution of best open weight model performance (\texttt{Command-A}) on different task clusters across African languages in downstream data.}
    \label{appdx_fig:performance_analysis_scatterplot}
\end{figure*}

\begin{figure*}[!ht]
  \centering
  \includegraphics[width=\textwidth]{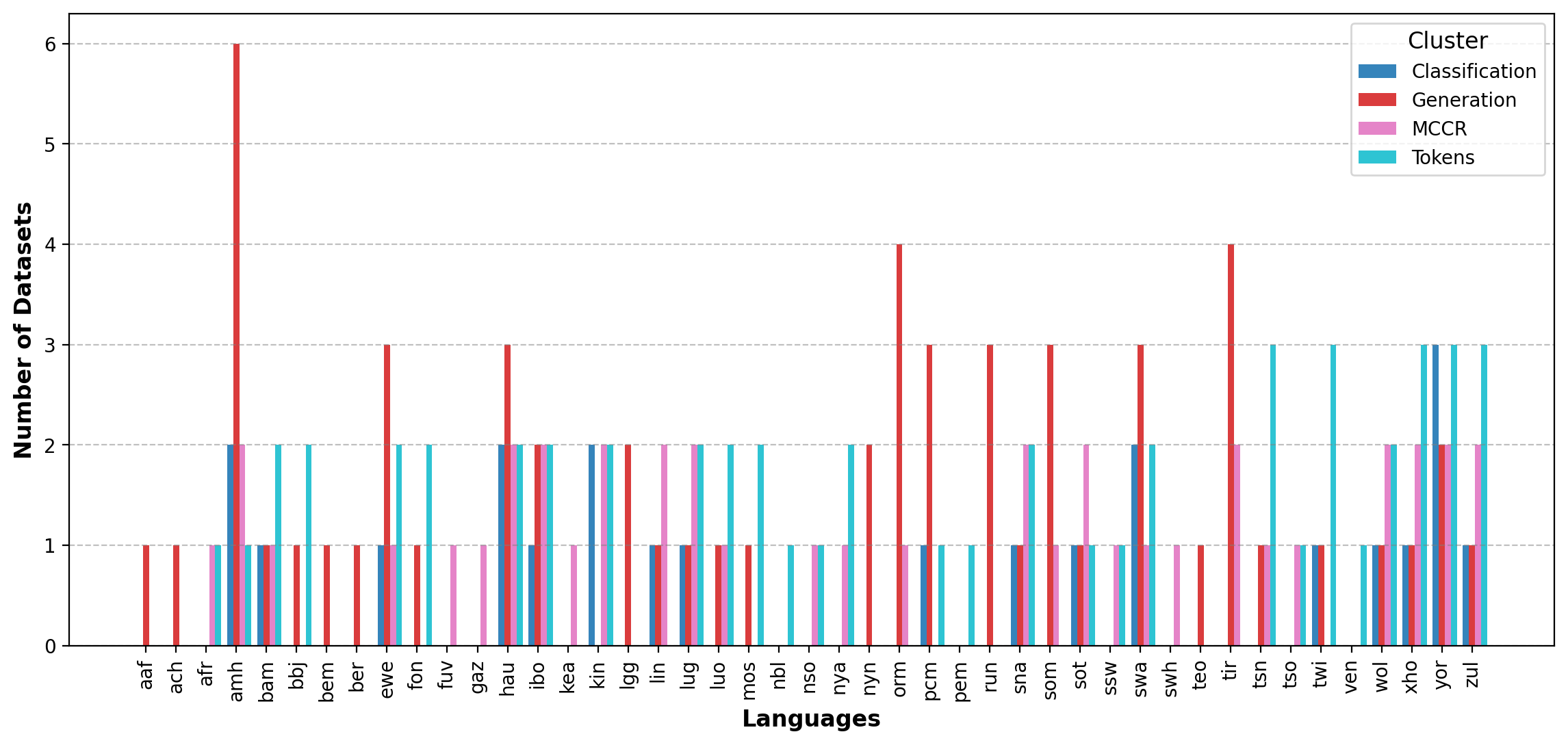}
  \caption{Distribution of languages across different clusters. Each bar represents the number of datasets a language appears in, categorized into the clusters,  with some languages appearing in all four clusters while others are only present in one or two.}   
  \label{fig:cluster_lang}
\end{figure*}

\begin{figure*}[!ht]
    \centering
    \resizebox{0.99\textwidth}{!}{ 
        \begin{minipage}{\linewidth} 
            \centering
            \subfigure[Command-A performers.]{
                \includegraphics[width=0.75\linewidth]{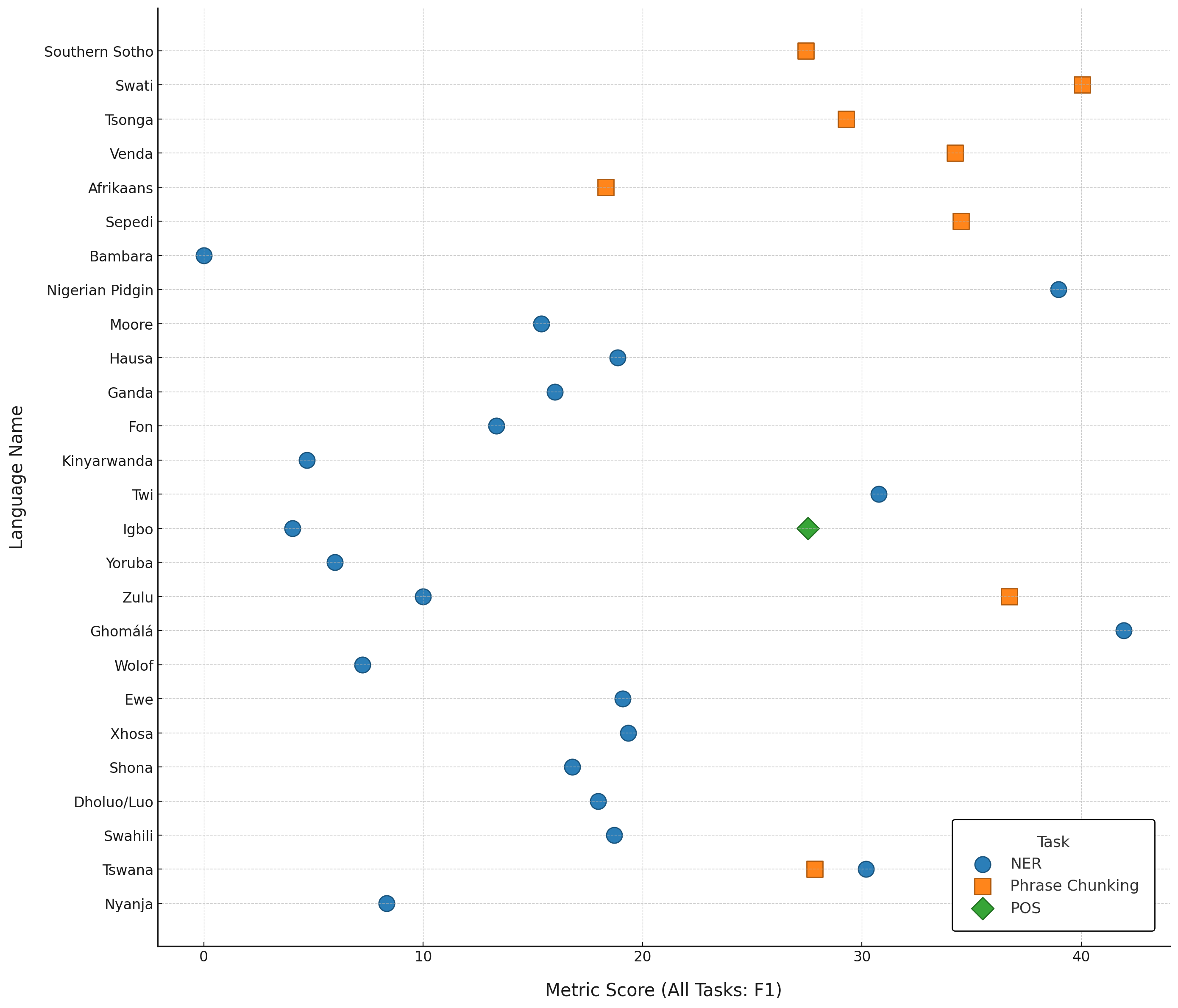}
                \label{fig:preform_sub1}
            }
            \subfigure[Claude-4-Sonnet performance.]{
                \includegraphics[width=0.75\linewidth]{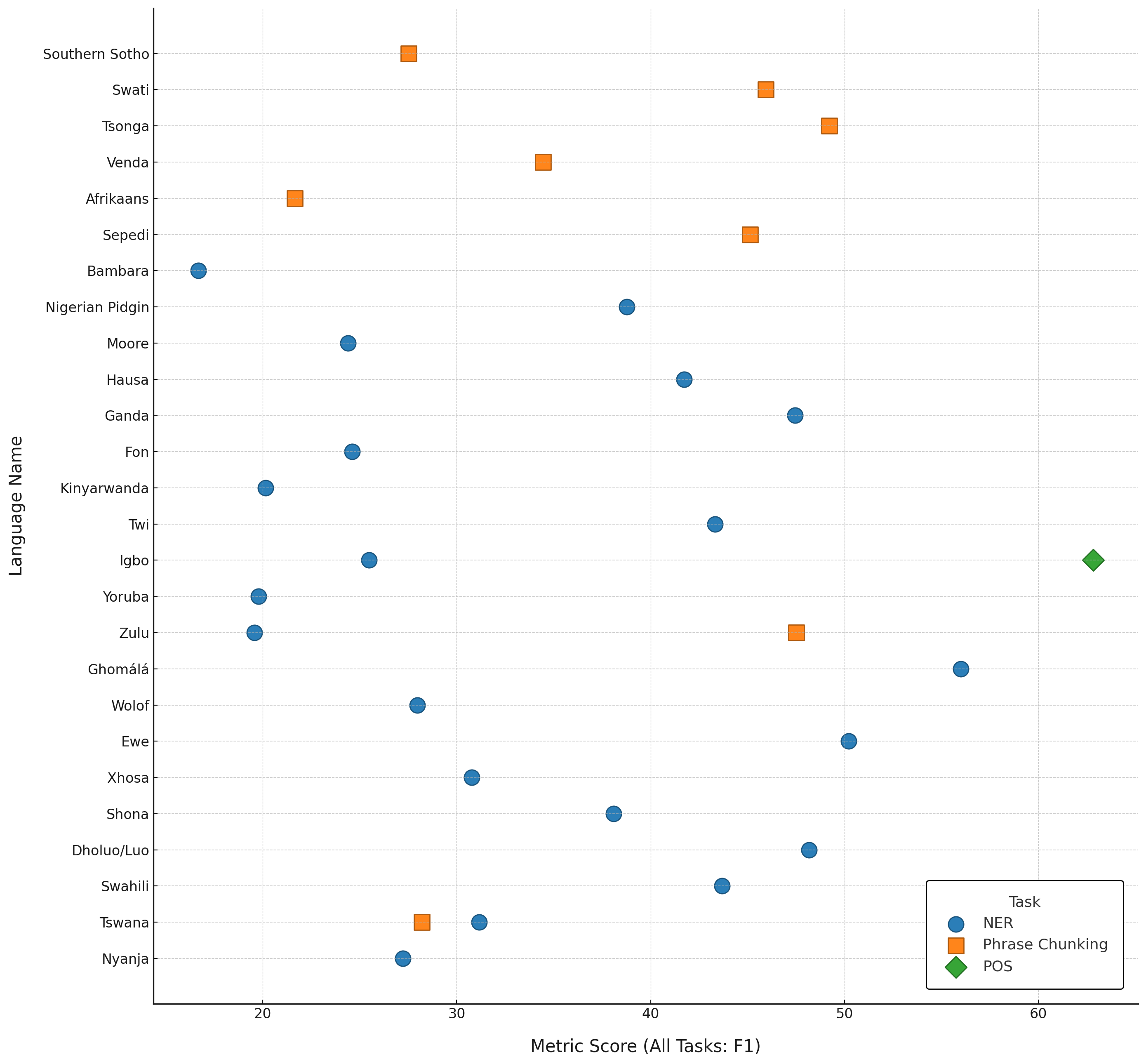}
                \label{fig:preform_sub3}
            }
            
        \end{minipage}
    }
    \caption{Performance of \texttt{Command-A} and \texttt{Claude-4-Sonnet} on token-level task cluster across African languages in downstream data.}
    \label{fig:apend_performance_analysis_tokens}
\end{figure*}

\begin{sidewaystable*}[]
\resizebox{\textwidth}{!}{%
\begin{tabular}{cllcrrrrrrlrrrrrrrrrrrrrrrrlrr}
\toprule
\multicolumn{1}{c}{}                                                & \multicolumn{1}{c}{}                                      & \multicolumn{1}{c}{}                                  & \multicolumn{1}{c}{}                                  & \multicolumn{6}{c}{\textbf{SLM}}                                                                                                                                                                          & \textbf{} & \multicolumn{16}{c}{\textbf{LLM}}                                                                                                                                                                                                                                                                                                                                                                                                                                                                                                                                                                                                                                                                                                                                                                                                               & \textbf{}                                                & \multicolumn{2}{c}{\textbf{Closed LLM}}                                                                                       \\ \cmidrule{5-10} \cmidrule{12-27} \cmidrule{29-30}
\multicolumn{1}{c}{\multirow{-2}{*}{\textbf{Cluster}}}              & \multicolumn{1}{c}{\multirow{-2}{*}{\textbf{Identifier}}} & \multicolumn{1}{c}{\multirow{-2}{*}{\textbf{Metric}}} & \multicolumn{1}{c}{\multirow{-2}{*}{\textbf{Shots}}} & \multicolumn{1}{c}{\textbf{\colorbox{yellow!20}{\textbf{S$_1$}}}} & \multicolumn{1}{c}{\textbf{\colorbox{yellow!20}{\textbf{S$_2$}}}} & \multicolumn{1}{c}{\textbf{\colorbox{yellow!20}{\textbf{S$_3$}}}} & \multicolumn{1}{c}{\textbf{\colorbox{yellow!20}{\textbf{S$_4$}}}} & \multicolumn{1}{c}{\textbf{\colorbox{yellow!20}{\textbf{S$_5$}}}} & \multicolumn{1}{c}{\textbf{\colorbox{yellow!20}{\textbf{S$_6$}}}} & \textbf{} & \multicolumn{1}{c}{\textbf{\colorbox{blue!20}{\textbf{L$_1$}}}} & \multicolumn{1}{c}{\textbf{\colorbox{blue!20}{\textbf{L$_2$}}}} & \multicolumn{1}{c}{\textbf{\colorbox{blue!20}{\textbf{L$_3$}}}} & \multicolumn{1}{c}{\textbf{\colorbox{blue!20}{\textbf{L$_4$}}}}   & \multicolumn{1}{c}{\textbf{\colorbox{blue!20}{\textbf{L$_5$}}}}   & \multicolumn{1}{c}{\textbf{\colorbox{blue!20}{\textbf{L$_6$}}}}                               & \multicolumn{1}{c}{\textbf{\colorbox{blue!20}{\textbf{L$_7$}}}}                               & \multicolumn{1}{c}{\textbf{\colorbox{blue!20}{\textbf{L$_8$}}}}                               & \multicolumn{1}{c}{\textbf{\colorbox{blue!20}{\textbf{L$_9$}}}} & \multicolumn{1}{c}{\textbf{\colorbox{blue!20}{\textbf{L$_{10}$}}}}                              & \multicolumn{1}{c}{\textbf{\colorbox{blue!20}{\textbf{L$_{11}$}}}}                              & \multicolumn{1}{c}{\textbf{\colorbox{blue!20}{\textbf{L$_{12}$}}}}                              & \multicolumn{1}{c}{\textbf{\colorbox{blue!20}{\textbf{L$_{13}$}}}} & \multicolumn{1}{c}{\textbf{\colorbox{blue!20}{\textbf{L$_{14}$}}}}                              & \multicolumn{1}{c}{\textbf{\colorbox{blue!20}{\textbf{L$_{15}$}}}}                              & \multicolumn{1}{c}{\textbf{\colorbox{blue!20}{\textbf{L$_{16}$}}}}                              & \textbf{}                                                & \multicolumn{1}{c}{\textbf{\colorbox{red!20}{\textbf{CL$_1$}}}}                              & \multicolumn{1}{c}{\textbf{\colorbox{red!20}{\textbf{CL$_2$}}}}                             \\ \midrule
 \multirow{6}{*}{\rotatebox[origin=c]{90}{\textbf{Classification}}} & xlni & Acc. & 5 & 33.6 & 33.7 & 36.7 & 41.18 & 40.4 & 46.68 &  & 48.08 & 55.06 & 34.3 & 40.69 & 37.09 & 61.09 & 57.86 & \underline{65.58} & 44.2 & 38.19 & 51.89 & 48.38 & 52.09 & 35.87 & 46.77 & 59.67 &  & 69 & \colorbox{green!20}{\textbf{69.6}}\\
 & lid & $F_1$ &10& 0 & 0 & 0.87 & 0.26 & 0.85 & 0.77 &  & 2.82 & 3.6 & 0 & 1.49 & 2.27 & 1.95 & \colorbox{green!20}{\underline{\textbf{4.61}}} & 4 & 1.81 & 3.34 & 3.26 & 3.07 & 3.3 & 3.05 & 3.32 & 4.57 &  & 3.73 & 4.48\\
 & news & $F_1$ & 3 & 0.86 & 0.35 & 0.15 & 17.54 & 0 & 6.64 &  & 21.94 & 0 & 2.06 & 24.06 & 0.12 & 25.83 & 7.83 & \underline{35.56} & 28.03 & 0.61 & 35.26 & 33.31 & 34.81 & 3.45 & 30.46 & 28.66 &  & \colorbox{green!20}{\textbf{40.09}} & 32.19\\
 & sentiment & $F_1$ & 5 & 38.51 & 24.75 & 30.45 & 11.05 & 28.77 & 1.99 &  & 36.09 & 42.47 & 38.4 & 28.36 & 36.07 & 34.56 & 20.55 & 46.35 & 35.86 & 25.94 & 24.51 & 45.82 & 35.89 & 19.98 & 37.07 & \underline{47.14} &  & 55.67 & \colorbox{green!20}{\textbf{57.52}}\\
 & topic & $F_1$ & 3 & 8.73 & 31.62 & 20.3 & 19.79 & 12.46 & 17.15 &  & 48.11 & 39.35 & 34.95 & 28.55 & 42.12 & 44.3 & 31.8 & \underline{70.73} & 36.47 & 43.74 & 64.89 & 56.48 & 48.73 & 52.09 & 50.96 & 53.16 &  & 67.93 & \colorbox{green!20}{\textbf{76.57}}\\ \cmidrule{3-30}
 &  & \textbf{Avg.} &  & 16.34 & 18.08 & 17.69 & 17.96 & 16.50 & 14.65 &  & 31.41 & 28.10 & 21.94 & 24.63 & 23.53 & 33.55 & 24.53 & \underline{44.44} & 29.27 & 22.36 & 35.96 & 37.41 & 34.96 & 22.89 & 33.72 & 38.64 &  & 47.28 & \colorbox{orange!20}{\textbf{48.07}}\\ \midrule
\multirow{7}{*}{\rotatebox[origin=c]{90}{\textbf{Generation}}}  & mt\_eng2xx & spBleu\textsuperscript{1K} & 5 & 1.79 & 3.63 & 3.45 & 3.25 & 2.12 & 3.4 &  & 7.07 & 7.61 & 1.72 & 4.37 & 2.79 & 6.24 & 6.91 & 4.56 & 3.73 & 3.29 & 8.84 & 8.7 & \underline{9.36} & 1.4 & 6.06 & 8.22 &  & 12.18 & \colorbox{green!20}{\textbf{12.36}}\\
 & mt\_fra2xx & spBleu\textsuperscript{1K} & 5 & 0.79 & 0.36 & 0.74 & 1.05 & 1.25 & 1.32 &  & 0.73 & 0.95 & 0.7 & 1.19 & 0.66 & 2.03 & 1.67 & 2.05 & 1.15 & 0.82 & 1.93 & 1.87 & 1.77 & 1.05 & 1.04 & \underline{2.3} &  & 3.41 & \colorbox{green!20}{\textbf{3.42}}\\
 & mt\_xx2xx & spBleu\textsuperscript{1K} & 5 & 0.4 & 0.45 & 0.42 & 0.42 & 0.36 & 0.18 &  & 0.43 & 0.63 & 0.72 & 0.55 & 0.46 & 0.71 & 0.9 & \underline{1.39} & 0.55 & 0.17 & 1.28 & 1.05 & 1.37 & 0.67 & 0.38 & 0.85 &  & \colorbox{green!20}{\textbf{5.15}} & 4.44\\
 & paraphrase & spBleu\textsuperscript{1K} & 5 & 23.2 & 28.88 & 31.15 & 19.12 & 25.93 & 17.85 &  & 28.73 & 25.83 & 31.68 & 31.33 & 23.66 & 17.26 & 21.34 & 15.37 & 32.13 & 30.68 & 26.86 & 19.64 & 27.79 & 19.23 & \colorbox{green!20}{\underline{\textbf{36.06}}} & 26.06 &  & 23.2 & 20.35\\
 & summary & rougeL & 2 & 10.03 & 3.24 & 4.37 & 10.07 & 0.91 & 13.34 &  & 15.1 & 10.43 & 1.45 & 12.74 & 5.92 & 16.16 & 15.73 & 17.13 & 13.27 & 4.19 & 16.92 & \colorbox{green!20}{\underline{\textbf{17.19}}} & 16.4 & 15.46 & 14.88 & 16.09 &  & 5.99 & 16.3\\
 & title & spBleu\textsuperscript{1K} & 2 & 2.14 & 2.81 & 0.25 & 2.3 & 0.03 & 5.78 &  & 7.48 & 5.78 & 3 & 5.94 & 1.58 & 7.75 & 7.51 & 8.65 & 6.83 & 3.2 & \underline{11.07} & 10.34 & 8.78 & 7.32 & 6.69 & 8.65 &  & \colorbox{green!20}{\textbf{13.62}} & 9.48\\ \cmidrule{3-30}
 &  & \textbf{Avg.} &  & 6.39 & 6.56 & 6.73 & 6.04 & 5.10 & 6.98 &  & 9.92 & 8.54 & 6.55 & 9.35 & 5.85 & 8.36 & 9.01 & 8.19 & 9.61 & 7.06 & \colorbox{orange!20}{\underline{\textbf{11.15}}} & 9.80 & 10.91 & 7.52 & 10.85 & 10.36 &  & 10.59 & 11.06\\  \midrule
\multirow{5}{*}{\rotatebox[origin=c]{90}{\textbf{MCCR}}} & mmlu & Acc. & 5 & 24.09 & 28.5 & 28.68 & 27.29 & 28.9 & 30.99 &  & 31.88 & 40.9 & 26.4 & 44.51 & 22.1 & 42.09 & 44.9 & 60.71 & 51.5 & 24 & 56.51 & 59.3 & 50.42 & 59.81 & 29.8 & \colorbox{green!20}{\underline{\textbf{61.91}}} &  & 81.6 & 58.6\\
 & mgsm & ExactM & 5 & 2.6 & 3.21 & 0 & 1.2 & 3.39 & 0.7 &  & 2.9 & 2.8 & 3.3 & 2.7 & 1.5 & 2.3 & 5.3 & 0.6 & 4.4 & 3 & 10.1 & 7.51 & 7.6 & 4.5 & 4.39 & \underline{11.1} &  & \colorbox{green!20}{\textbf{45.9}} & 29.5\\
 & belebele & Acc. & 5 & 27.71 & 14.71 & 25.2 & 29.31 & 31.31 & 24.69 &  & 22.7 & 26.93 & 25.09 & 28.24 & 26.33 & 25.63 & 31.04 & 32.34 & 33.64 & 26.61 & 30.4 & \underline{35.92} & 33.72 & 30.83 & 20.99 & 29.12 &  & 36.6 & \colorbox{green!20}{\textbf{39.7}}\\
 & squad\_qa & $F_1$ & 5 & 57.45 & 70.89 & 67.32 & 43.13 & 71.31 & 74.97 &  & 78.04 & 80.91 & 52.66 & 59.48 & 63.41 & 73.78 & \colorbox{green!20}{\underline{\textbf{82.49}}} & 79.16 & 69.25 & 64.64 & 77.65 & 76.35 & 66.25 & 53.92 & 74.64 & 80.05 &  & 78.02 & 76.11\\ \cmidrule{3-30}
 &  & \textbf{Avg.} &  & 27.96 & 29.33 & 30.30 & 25.23 & 33.73 & 32.84 &  & 33.88 & 37.89 & 26.86 & 33.73 & 28.34 & 35.95 & 40.93 & 43.20 & 39.70 & 29.56 & 43.67 & 44.77 & 39.50 & 37.27 & 32.46 & \underline{45.55} &  & \colorbox{orange!20}{\textbf{60.53}} & 50.98\\  \midrule
\multirow{4}{*}{\rotatebox[origin=c]{90}{\textbf{Tokens}}} & ner & $F_1$ & 5 & 0 & 0.78 & 1.76 & 2.99 & 0.13 & 3.65 &  & 2.26 & 7.49 & 2.35 & 3.96 & 0.2 & 6.03 & 13.14 & 11.4 & 8.51 & 1.52 & 6.86 & 10.54 & 8.64 & 6.04 & 4.3 & \underline{16.88} &  & \colorbox{green!20}{\textbf{35.06}} & 27.04\\
 & phrase & $F_1$ & 5 & 14.79 & 28.5 & 29.44 & 11.52 & 24.91 & 27.73 &  & 15.95 & 29.35 & 29.42 & 16.55 & 30.87 & 13.59 & 30.01 & 23.97 & 26.34 & 23.09 & 25.55 & 23.13 & 16.78 & 25.13 & 29.44 & \underline{31.05} &  & \colorbox{green!20}{\textbf{36.69}} & 36.45\\
 & pos & $F_1$ & 5 & 8.83 & 6.73 & 10.56 & 8.04 & 10.3 & 10.46 &  & 9.13 & 13.2 & 9.81 & 11.68 & 12.52 & 13.86 & 19 & 13.99 & 18.36 & 11.4 & 14.13 & 15.04 & 11.52 & 20.76 & 13.71 & \underline{27.54} &  & \colorbox{green!20}{\textbf{62.84}} & 38.67\\ \cmidrule{3-30}
 &  & \textbf{Avg.} &  & 7.87 & 12.00 & 13.92 & 7.52 & 11.78 & 13.95 &  & 9.11 & 16.68 & 13.86 & 10.73 & 14.53 & 11.16 & 20.72 & 16.45 & 17.74 & 12.00 & 15.51 & 16.24 & 12.31 & 17.31 & 15.82 & \underline{25.16} &  & \colorbox{orange!20}{\textbf{44.86}} & 34.05\\ \midrule
 &  & \textbf{Overall} &  & 14.64 & 16.49 & 17.16 & 14.19 & 16.78 & 17.10 &  & 21.08 & 22.80 & 17.30 & 19.61 & 18.06 & 22.25 & 23.80 & 28.07 & 24.08 & 17.75 & 26.57 & 27.05 & 24.42 & 21.25 & 23.21 & \underline{29.93} &  & \colorbox{orange!20}{\textbf{40.82}} & 36.04\\ \bottomrule
\end{tabular}%
}
\caption{Few-shot evaluation across different task clusters/tasks. The highest score for each individual task is in \colorbox{green!20}{\textbf{bold green}}, while the best average score across each task cluster (and \textbf{overall} score, calculated as average of task cluster averages) is in \colorbox{orange!20}{\textbf{bold orange}}. \underline{\textbf{Underline}}: refers to the best score across the open source/weights LLMs. \textbf{Shots}: number of shots. \textbf{SLMs:} \colorbox{yellow!20}{\textbf{S$_1$}} Llama3.2 (1B), \colorbox{yellow!20}{\textbf{S$_2$}} Llama3.2 (3B), \colorbox{yellow!20}{\textbf{S$_3$}} Gemma2 (2B), \colorbox{yellow!20}{\textbf{S$_4$}} Phi-3.5 (3.8B), \colorbox{yellow!20}{\textbf{S$_5$}} Phi-4 (3.8B), and \colorbox{yellow!20}{\textbf{S$_6$}} Gemma3 (4B). \textbf{LLMs:} \colorbox{blue!20}{\textbf{L$_1$}} Llama3.1 (8B), \colorbox{blue!20}{\textbf{L$_2$}} Gemma2 (9B), \colorbox{blue!20}{\textbf{L$_3$}} Aya (8B), \colorbox{blue!20}{\textbf{L$_4$}} Babel (9B), \colorbox{blue!20}{\textbf{L$_5$}} Command-R-Plus-R7B (8B), \colorbox{blue!20}{\textbf{L$_6$}} Gemma3 (12B), \colorbox{blue!20}{\textbf{L$_7$}} Gemma2 (27B), \colorbox{blue!20}{\textbf{L$_8$}} Gemma3 (27B), \colorbox{blue!20}{\textbf{L$_9$}} DeepSeek-R1-Distill-Qwen (32B), \colorbox{blue!20}{\textbf{L$_{10}$}} Aya (35B), \colorbox{blue!20}{\textbf{L$_{11}$}} Llama3.1 (70B), \colorbox{blue!20}{\textbf{L$_{12}$}} Llama3.3 (70B), \colorbox{blue!20}{\textbf{L$_{13}$}} DeepSeek-R1-Distill-Qwen (70B), \colorbox{blue!20}{\textbf{L$_{14}$}} Babel (83B), \colorbox{blue!20}{\textbf{L$_{15}$}} Command-R-Plus (104B), \colorbox{blue!20}{\textbf{L$_{16}$}} Command-A (111B). \textbf{CLosed LLMs:} \colorbox{red!20}{\textbf{CL$_1$}} Claude-4-Sonnet, \colorbox{red!20}{\textbf{CL$_2$}} GPT-4.1.}
\label{appd_tab:results-subset}
\end{sidewaystable*}

\end{document}